\documentclass[a4paper,fleqn]{cas-dc}
\usepackage[numbers,sort&compress]{natbib}
\usepackage{amssymb}
\usepackage{graphicx}
\usepackage{caption}
\usepackage{stmaryrd}
\usepackage{latexsym,amssymb}
\usepackage{amsmath,amsbsy}
\usepackage{amsopn,amstext}
\usepackage{bm}
\usepackage{float}
\usepackage{picins}
\usepackage{subfigure}
\usepackage{graphicx}
\usepackage{color}
\usepackage{longtable}
\usepackage{booktabs}
\usepackage{multirow}
\usepackage{graphicx}
\usepackage{threeparttable}
\usepackage{makecell}
\usepackage{tabularx}
\usepackage{pdfpages}

\usepackage{multirow}
\usepackage{algorithm}
\usepackage{algpseudocode}
\usepackage{hyperref}

\definecolor{hyperref-blue}{RGB}{0, 0, 0}
\newcommand{\reffig}[1]{Fig. \ref{#1}}

\newcommand{\reftable}[1]{Table \ref{#1}}

\def\tsc#1{\csdef{#1}{\textsc{\lowercase{#1}}\xspace}}
\tsc{WGM}
\tsc{QE}
\tsc{EP}
\tsc{PMS}
\tsc{BEC}
\tsc{DE}

\begin{document}
\let\WriteBookmarks\relax
\def\floatpagepagefraction{1}
\def\textpagefraction{.001}

\shorttitle{}    
\shortauthors{Junkai Mao et ~ al.} 
\title [mode = title]{Backbone-based Dynamic Graph Spatio-Temporal Network for Epidemic Forecasting}  
%
\author[1]{Junkai Mao}[type=editor ]
\ead{maojk@shu.edu.cn}
\credit{Investigation, Methodology, Modeling, Simulation, Validation, Writing - original draft}
%
%
\author[1,2,3]{Yuexing Han}[type=editor, orcid=0000-0002-1170-202X, ]
\cormark[1]
\ead{han_yx@i.shu.edu.cn}
\cortext[cor1]{Corresponding author}
\credit{Review, Editing}
\author[4,5]{Gouhei Tanaka}[type=editor, orcid=0000-0002-6223-4406, ]
\ead{gtanaka@nitech.ac.jp}
\credit{Review, Editing}
\author[1]{Bing Wang}[type=editor, orcid=0000-0002-7078-4352, ]
\cormark[2]
\ead{bingbignwang@shu.edu.cn}
\cortext[cor1]{Corresponding author}
\credit{Modeling, Validation, Review, Editing}
%
\address[1]{School of Computer Engineering and Science, Shanghai University, Shanghai 200444, PR China}
\address[2]{Key Laboratory of Silicate Cultural Relics Conservation (Shanghai University), Ministry of Education, PR China}
\address[3]{Zhejiang Laboratory, Hangzhou 311100, PR China}
\address[4]{International Research Center for Neurointelligence, The University of Tokyo, Tokyo 1130033, Japan}
\address[5]{Department of Computer Science, Graduate School of Engineering, Nagoya Institute of Technology, Nagoya 4668555, Japan}
\begin{abstract}
Accurate epidemic forecasting is a critical task in controlling disease transmission. Many deep learning-based models focus only on static or dynamic graphs when constructing spatial information, ignoring their relationship. Additionally, these models often rely on recurrent structures, which can lead to error accumulation and computational time consumption. To address the aforementioned problems, we propose a novel model called Backbone-based Dynamic Graph Spatio-Temporal Network (BDGSTN). Intuitively, the continuous and smooth changes in graph structure, make adjacent graph structures share a basic pattern. To capture this property, we use adaptive methods to generate static backbone graphs containing the primary information and temporal models to generate dynamic temporal graphs of epidemic data, fusing them to generate a backbone-based dynamic graph. To overcome potential limitations associated with recurrent structures, we introduce a linear model DLinear to handle temporal dependencies and combine it with dynamic graph convolution for epidemic forecasting. Extensive experiments on two datasets demonstrate that BDGSTN outperforms baseline models and ablation comparison further verifies the effectiveness of model components. Furthermore, we analyze and measure the significance of backbone and temporal graphs by using information metrics from different aspects. Finally, we compare model parameter volume and training time to confirm the superior complexity and efficiency of BDGSTN.
\end{abstract} 

\maketitle

\section{Introduction}
\par In the past few years, the rapid global spread of the COVID-19, due to its high transmissibility and relatively high morbidity rate, has had a significant impact on the global economy, trade, healthcare resources, and human lives~\cite{kaye2021economic}. This global epidemic has posed enormous challenges to many countries and regions, compelling them to adopt a series of measures to suppress the epidemic spread, including social distancing, face mask, and vaccination~\cite{kwon2021association}. Accurate forecasting of epidemic spread is crucial for assisting policymakers in formulating effective strategies for epidemic control, allocating resources sensibly, and enhancing healthcare provision to safeguard human lives. However, epidemic evolution is a complex real-world phenomenon, and simple mathematical models often struggle to capture the underlying nonlinear dynamics of an epidemic. Fortunately, with the rapid advancements of deep learning in various fields like computer vision, natural language processing, and data mining, researchers have begun applying deep learning techniques to epidemic forecasting~\cite{wang2021improving,ning2023epi}. They accomplish this by constructing sophisticated deep neural networks to assist in capturing the dynamics of epidemic transmission, thereby improving the accuracy of epidemic forecasting.
\par Early researchers proposed a series of compartmental models based on differential equations to simulate the spread of epidemics among different populations. Among them, the SIR model~\cite{sattenspiel1995structured} and SEIR model~\cite{efimov2021interval} are considered as fundamental compartmental models. By utilizing differential equations that represent the temporal changes in the populations, the models can well fit the dynamics of epidemic transmission. Based on these models, researchers have developed various variant models to incorporate the characteristics of actual transmission and intervention scenarios such as vaccination~\cite{turkyilmazoglu2022extended}, asymptomatic infections~\cite{de2020seiard}, and transmission during the latent period~\cite{lopez2021modified}. By incorporating these additional factors, researchers can better capture the nature of real-world epidemic transmission and improve the accuracy of predictions. In addition, epidemic forecasting can be classified as a time series forecasting task due to its temporal characteristics. Many traditional time series analysis methods can be directly applied to epidemic forecasting, such as ARIMA~\cite{hernandez2020forecasting} and SVR~\cite{parbat2020python} models. These traditional methods model and forecast historically confirmed cases, contributing to a better understanding of the dynamic transmission characteristics of epidemics~\cite{singh2020study}. Several representative and effective deep learning models have been proposed, including recurrent networks such as RNN~\cite{connor1994recurrent}, LSTM~\cite{hochreiter1997long}, and GRU~\cite{chung2014empirical}, convolutional networks like TCN~\cite{bai2018empirical}, differential equation-based models such as NODE~\cite{chen2018neural} and NCDE~\cite{kidger2020neural}, and a series of transformer-based variant models like Autoformer~\cite{wu2021autoformer}, FEDformer~\cite{zhou2022fedformer}, and PatchTST~\cite{nie2022time}. These models enable epidemic forecasting by learning the complex nonlinear temporal dynamics.
\par However, in reality, the spread of epidemics is not confined to a single patch but influenced by population movements between different patches~\cite{hazarie2021interplay}. As a result, epidemic data exhibits not only temporal dynamics but also spatial dependencies. This implies that relying solely on time series models may cause inaccuracies in epidemic forecasting. Therefore, epidemic forecasting can be regarded as a spatio-temporal forecasting problem, where historical data is modeled in both temporal and spatial dimensions to capture underlying spatio-temporal patterns. The emergence of Graph Convolutional Networks~\cite{kipfsemi} has provided researchers with effective tools for handling spatial data in a non-Euclidean space. Building upon this advancement, numerous spatio-temporal epidemic forecasting models have been proposed~\cite{deng2020cola,gao2021stan,wang2022causalgnn,cao2023mepognn,mao2023mpstan}, where most of them utilize recurrent structures based on RNNs. Furthermore, due to the similarities in spatio-temporal attributes, models developed for other spatio-temporal tasks, such as traffic flow forecasting and metro passenger flow forecasting, can also be applied to epidemic forecasting~\cite{zhao2019t,jiang2022deep}. These models utilize time-series models to capture temporal dependencies and employ graph algorithms to capture spatial dependencies. Among them, the accurate construction of graphs representing interactions between regions or patches is crucial for better forecasting accuracy since it controls the pathways of information aggregation within the graph~\cite{huang2023multi}. 
\par Most spatio-temporal forecasting models utilize prior knowledge to construct pre-defined and fixed graph structures. For example,~\cite{zhao2019t} utilizes geographical adjacency to construct a binary transportation network. However, in reality, the influences between different regions or patches are more diverse. Therefore,~\cite{wang2020traffic} further calculates a weight matrix based on adjacency to more accurately reflect the degree of influence. Additionally,~\cite{wang2022model} uses population mobility data to build an interaction network of patches, while~\cite{chen2021graph} aims to integrate more prior knowledge, such as adjacency, population mobility, and travel distance, to construct an informative metro relationship graph.  Furthermore,~\cite{gao2021stan,mao2023mpstan} employ the gravity model based on population and distance to calculate the mutual influences between different patches. In addition to prior knowledge, some studies utilize similarity algorithms such as DTW or PCC to calculate the similarity of time series data, thereby assisting in the construction of graph structures~\cite{liu2020physical,jiao2021graph,khaled2022tfgan}. Although the aforementioned methods are simple, the graphs generated based on prior knowledge or data similarity often exhibit intuitiveness, incompleteness, and biases, which do not accurately reflect the true relationships in the graph. Therefore,~\cite{wu2019graph,bai2020adaptive} propose the concept of adaptive adjacency matrix, which utilizes learnable embeddings to adaptively capture the underlying relationships. Additionally,~\cite{kong2022jointgraph} further leverages information propagation among nodes to compute node embeddings.
\par Real-world networks often exhibit dynamic characteristics~\cite{barros2021survey}, such as the time-varying population flow and interactions among different patches. Traditional static graph structures are inadequate in capturing the dynamic changes in these networks. In response to this issue, extensive research has attempted to take graph information as temporal data and apply temporal models to construct dynamic graphs that can adapt to network changes, thereby providing a more accurate representation of the graph information~\cite{li2022spatial,hu2022dstgcn}.~\cite{xu2023dynamic} leverages real-time dynamic data to assist in constructing dynamic graphs, but the application becomes challengeable in the absence of prior data. Additionally,~\cite{han2021dynamic} utilizes the inherent patterns in the data to generate dynamic graphs, specifically leveraging the notable periodicity observed in traffic flow data. Furthermore,~\cite{yin2023spatiotemporal,weng2023decomposition} incorporate static node embedding information into the construction of dynamic graphs to capture meaningful dynamics. In order to capture spatial dependencies comprehensively, some researchers perform graph convolution operations separately on static graphs and dynamic graphs, and aggregate the two convolutional information through weighted summation~\cite{li2023dynamic1,li2023dynamic2,kong2023dynamic}.

\par Although existing methods have achieved success in this field, we find several problems as follows, (1) Existing methods focus on constructing either static or dynamic graphs, overlooking the relationship between them. The limitation of this kind of approaches is that only constructing static graphs, thereby failing to capture the dynamic changes in the real graph. On the other hand, directly generating dynamic graphs may face challenges in terms of high complexity and computational costs, making it difficult to be effectively optimized by the loss function. Additionally, the generated dynamic graphs may exhibit significant differences between adjacent time steps. However,~\cite{yang2021discrete,ye2022learning} found that structural changes in real-world graphs are typically continuous and smooth rather than abrupt. Therefore, dynamic graphs with substantial differences may fail to accurately capture the graph evolution information in the real world. (2) Most spatio-temporal epidemic models commonly employ recurrent structures to handle the temporal dependencies. However, recurrent structures themselves present some issues. Firstly, they are susceptible to problems like gradient explosion or gradient vanishing, which become more pronounced when dealing with long sequences. Secondly, recurrent structures tend to accumulate errors when handling epidemic data with significant noise. Additionally, the iterative nature of recurrent structures for information propagation can result in longer computation time when dealing with large-scale data.
\par To address the above issues, we propose a method named Backbone-based Dynamic Graph Spatio-Temporal Network (BDGSTN) for epidemic forecasting. Taking into account the continuous and smooth changes in graph structures, we assume that adjacent graph structures are similar or share a basic graph pattern. Unlike previous methods that construct static or dynamic graphs, we separately construct a static backbone network and a dynamic temporal network. The former network captures shared graph structure information over a period of time using an adaptive approach, while the latter learns the time-series information of the graph structures using a temporal model. By integrating the backbone and temporal networks, we establish an integrated graph network that represents the dynamic interactions of patches. Furthermore, inspired by~\cite{zeng2022transformers}, we recognize that simple linear models outperform other temporal models in terms of prediction accuracy and computational efficiency for temporal data forecasting. To avoid the issues associated with recurrent structures, we employ a linear model called DLinear based on time series decomposition to handle temporal dependencies and combine GCN with the dynamic graph to address spatial dependencies, enabling spatio-temporal epidemic forecasting.  In summary, the main contributions of this paper are as follows:
\begin{figure*}[h]
\centering
\includegraphics[width=\linewidth]{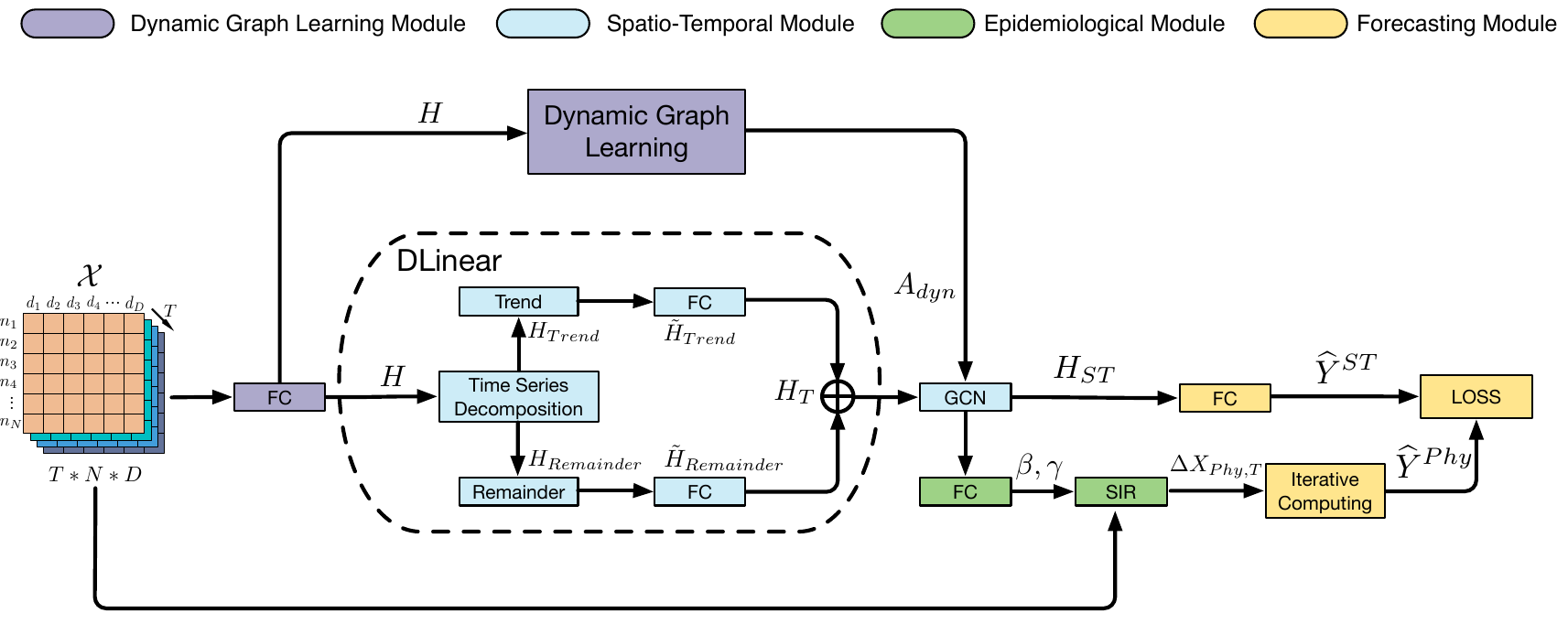}
\caption{The framework of the BDGSTN model.}
\label{fig:BDGSTN}
\end{figure*}
\begin{enumerate}[(1)]
  \item We design a method for generating dynamic graphs based on the backbone graph. This method utilizes an adaptive technique to learn the backbone graph and employs a temporal model to capture the temporal graphs of epidemic data. The backbone graph captures the shared structural information of the dynamic graph, while the temporal graphs represent the time-series information or dynamic changes. By integrating these two types of graph information, we further construct a more representative dynamic graph.
  \item We propose a novel backbone-based dynamic graph spatio-temporal learning model named BDGSTN for epidemic forecasting. This network combines DLinear, which utilizes time-series decomposition, and GCN, which leverages the dynamic graph, to effectively address the underlying spatio-temporal dependencies of epidemics, thus achieving accurate forecasting.
  \item We conduct extensive experiments to validate the performance of BDGSTN on two different datasets. The experimental results demonstrate that BDGSTN achieves State-of-the-Art or competitive accuracy in both short-term and long-term forecasting. Additionally, we further verify the effectiveness of the backbone-based dynamic graph and explore the roles of the backbone graph and temporal graph in the dynamic graph using information metrics.
\end{enumerate}  
\par The remainder of this paper is structured as follows: Section \ref{sec:Methodologies} provides a detailed description of the proposed model structure. Section \ref{sec:Experiments} presents the experimental results and further analyzes the findings. Finally, we summarize the entire work and outlook the future work in Section \ref{sec:Conclusion}.




\begin{figure*}[h]
\centering
\includegraphics[width=\linewidth]{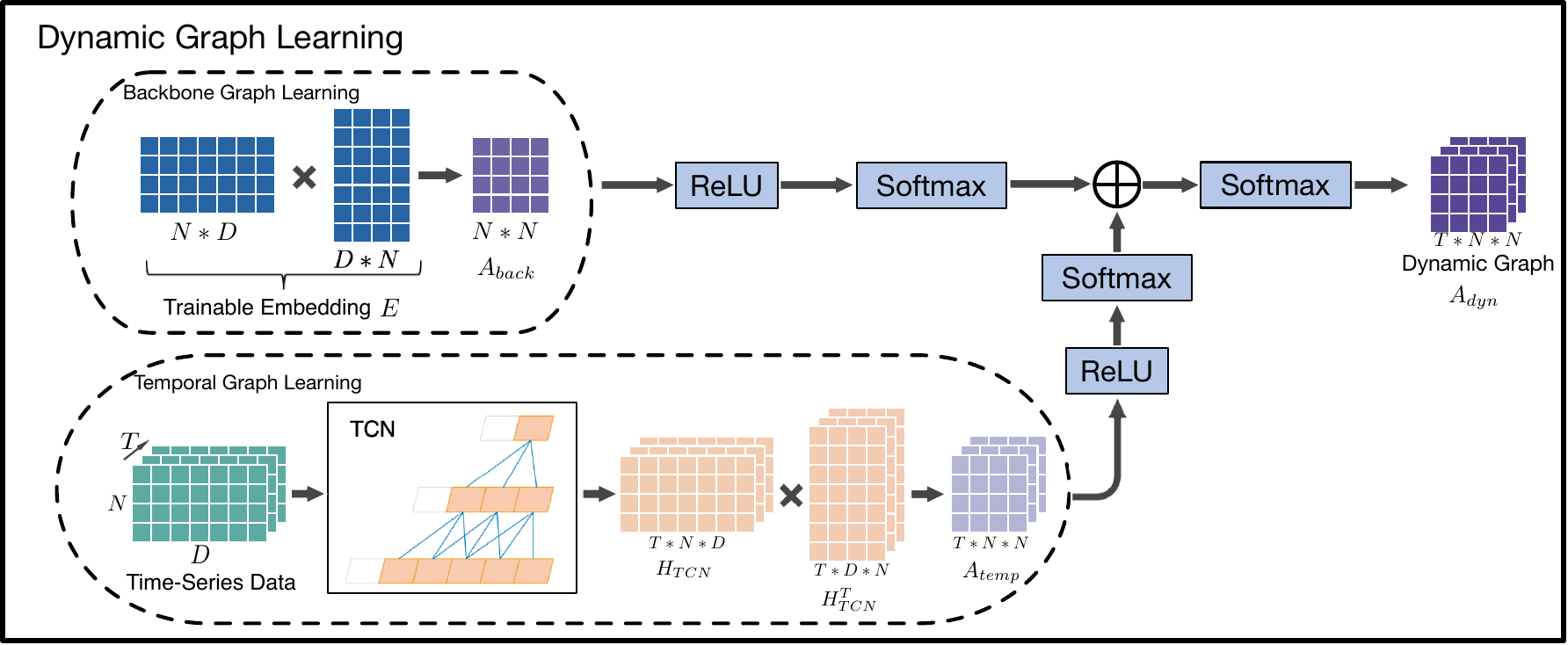}
\caption{The framework of the dynamic graph learning.}
\label{fig:dynamic}
\end{figure*}

\section{Methodology} \label{sec:Methodologies}
In this section, we first define the problem of spatio-temporal epidemic forecasting. Then, we present the overall framework of the proposed model and discuss the specific details of each module.
\subsection{Problem Definition}
\par The multi-patch epidemic network can be represented using a graph $G(\mathcal{V} ,\mathcal{E} )$, where $\mathcal{V}$ represents the set of patches and $\mathcal{E}$ represents the set of edges between patches. The graph $G$ can be converted into an adjacency matrix $A \in \mathbb{R} ^{N\times N}$, where $A_{ij}$ represents the connection weight between patch $i$ and $j$, and $N$ represents the number of patches. Since the network is time-varying, the adjacency matrix $A$ will also change accordingly. We use $\mathcal{A}_{1:T}=[A_{1},A_{2},\dots,A_{T}]\in \mathbb{R} ^{T\times N\times N}$ to denote the dynamic adjacency matrix over a specific time period.
\par We use $\mathcal{X}_{1:T}=[X_{1},X_{2},\dots,X_{T}]\in \mathbb{R} ^{T\times N\times D}$ to represent the spatio-temporal features of epidemic data, where $X_{t}$ with $t\in \left [ 1,T \right ] $ represents the $D$-dimensional historical data of $N$ patches at time step $t$, including daily infected cases, daily recovered cases, and daily susceptible cases. The objective of spatio-temporal epidemic forecasting is to leverage spatio-temporal features $\mathcal{X}$ and their corresponding dynamic adjacency matrix $\mathcal{A}$ from historical $T$ time steps to learn a mapping function $\mathcal{F}$, to forecast daily infected cases $\hat{Y}_{T+1:T+L}\in \mathbb{R} ^{T\times N}$ for $N$ patches at the future $L$ time steps. Therefore, the problem can be formulated as follows:
\begin{align}
    \hat{Y}_{T+1:T+L}=\mathcal{F}(\mathcal{X}_{1:T},\mathcal{A}_{1:T}).
\end{align}
\subsection{Model Overview}
\par The overall framework of the BDGSTN model, as shown in \reffig{fig:BDGSTN}, consists of four modules: Dynamic Graph Learning Module, Spatio-Temporal Module, Epidemiological Module, and Forecasting Module. The Dynamic Graph Learning Module learns the underlying backbone graph information and temporal graph information. A representative dynamic graph is generated by fusing these two types of graph information. The generated dynamic graph and historical data are then passed to the Spatio-Temporal Module, which utilizes DLinear and GCN to capture the dynamic temporal information and spatial influences. The Epidemiological Module incorporates epidemiological domain knowledge to assist the model in better capturing the underlying dynamics of epidemics. Finally, in the Forecasting Module, the model integrates neural network forecasting with epidemiological knowledge to jointly constrain the model, resulting in enhanced accuracy.

\subsection{Dynamic Graph Learning Module}
\par The interactive graph based on the epidemic is crucial for spatio-temporal epidemic forecasting as it influences the information transmission among patches in spatial dimensions. In the real world, interactive graphs are time-varying, and static and fixed graph structures cannot meet the requirements. Furthermore, directly generating dynamic graphs can be computationally complex, difficult to optimize, and lack constraints, leading to significant differences between adjacent time steps.~\cite{yang2021discrete,ye2022learning} point out that graph structure changes are typically continuous and smooth, and thereby, dynamic graphs with significant differences may not reflect the real dynamic evolution accurately. Inspired by this view, we consider the nature of similar structures in graphs at adjacent time steps, leading us to propose a novel method for dynamic graph learning.
\par We consider the shared graph structure information existing in dynamic graphs as backbone graphs, while converting the temporal change information into temporal graphs. Then, we fuse these two types of graph information to generate a dynamic graph. In this process, we increase the constraint of dynamic graphs by leveraging graph structure information to better reflect the dynamic evolution of epidemics. The details are shown in \reffig{fig:dynamic}.
\par Adaptive learning graph structures have been widely used in the field of graph learning and can provide accurate static graph information. Based on this perspective, we utilize the trainable embedding matrix $E\in \mathbb{R} ^{ N\times D_{ada}}$ and its transpose $E^{T}\in \mathbb{R} ^{ D_{ada}\times N}$, where $D_{ada}$ denotes the matrix feature number, to generate the backbone graph $A_{back}\in \mathbb{R} ^{ N\times N}$ given by,
\begin{align}
    A_{back}=EE^{T}.
\end{align}%
\par To learn the dynamic changes of a graph over time, we begin by mapping the epidemic data $X\in \mathbb{R} ^{ N\times T\times D}$ to a high-dimensional time-series embedding $H\in \mathbb{R} ^{ N\times T\times D_{H}}$ using a fully connected (FC) layer, where $D$ and $D_{H}$ represent the dimensions before and after the mapping, respectively.The mapping can be written as follows: 
\begin{align}
\label{equ:map}
    H=FC(X).
\end{align}%
\par This embedding is then fed into a temporal model to learn temporal features, while the embedding is also passed through the spatio-temporal module for further processing.

\par Temporal Convolutional Network (TCN) is a model specifically designed to capture the causal relationships in temporal data and is particularly effective for learning dynamic time-series embedding. \reffig{fig:dynamic} further illustrates how TCN reflects the impact of past events on future events. Given an input sequence $H_{i}\in \mathbb{R} ^{ T\times D_{H}}$ of patch $i$, the TCN is defined as follows:
\begin{align}
    TCN(H_{i})&=H_{i}\star f_{1\times r}(t),\nonumber\\
    &=\sum_{s=0}^{r-1} f_{1\times r}(s) H_{i}(t-d\times s),
\end{align}%
where $f_{1\times r}$ denotes a convolution filter with kernel size $r$, $d$ indicates the dilated factor, and $TCN(H_{i})\in \mathbb{R} ^{ T\times D_{TCN}}$ as the output.
\par Similar to the backbone graph, the learned time-series embedding $H_{TCN}\in \mathbb{R} ^{ T\times N\times D_{TCN}}$ and its transpose $H^{T}_{TCN}\in \mathbb{R} ^{ T\times D_{TCN}\times N}$ multiplied together represent the temporal graph $A_{temp}\in \mathbb{R} ^{ T\times N\times N}$ containing dynamic information, calculated as follows:
\begin{align}
    &H_{TCN}=TCN(H),\\
    &A_{temp}=H_{TCN}H_{TCN}^{T}.
\end{align}%
\par To generate a dynamic graph by integrating the backbone graph and the temporal graph, we firstly apply activation functions such as ReLU and Softmax to each of them, and then fuse the two types of graphs by element-wise addition.Since the dynamic graph represents the influence weights between patches, we finally utilize the Softmax function to map it into the range of 0 to 1, generating the dynamic graph $A_{dyn}\in \mathbb{R} ^{ T\times N\times N}$. The calculations are as follows:
\begin{align}
    &A_{back}=Softmax(ReLU(A_{back})),\\
    &A_{temp}=Softmax(ReLU(A_{temp})),\\
    &A_{dyn}=Softmax(A_{back}\oplus A_{temp}).
\end{align}%
\subsection{Spatio-Temporal Module}
\par The spatio-temporal module is used to model the spatio-temporal attributes of epidemic data. As suggested by~\cite{zeng2022transformers}, simple linear models may be more efficient and accurate than complex models when dealing with time series data. Therefore, we choose the DLinear model, which considers the trend evolution of the data, to capture temporal dependencies, while avoiding problems such as the accumulation of errors in the recurrent structure. Additionally, we employ the widely used GCN model in conjunction with the generated dynamic graph to capture dynamic spatial dependencies.
\par The DLinear model utilizes a moving average kernel to decompose the input data into trend and remainder components, a technique widely applied in temporal models~\cite{wu2021autoformer,zhou2022fedformer}. Firstly, we employ a moving average with a pooling operation to extract the trend component $H_{Trend}\in \mathbb{R} ^{ N\times T\times D_{H}}$. Then, we calculate the remainder component $H_{Remainder}\in \mathbb{R} ^{ N\times T\times D_{H}}$ by computing the difference between the input data and the trend. The calculation formulas are as follows:
\begin{align}
    &H_{Trend}=AvgPool(Padding(H)),\\
    &H_{Remainder}=H-H_{Trend},\\
    &H_{Trend},H_{Remainder}= SeriesDecomp(H),
\end{align}%
where $AvgPool()$ denotes the moving average with padding operation and $SeriesDecomp()$ is utilized to summarize the process of time series decomposition.

\par The extracted trend and residual components are linearly transformed using fully connected layers to generate new temporal embedding representations $\tilde{H}_{Trend}, \tilde{H}_{Remainder}\in \mathbb{R} ^{ N\times T\times D_{H}}$ for the trend and remainder. Subsequently, the updated trend and remainder components are aggregated using addition operation to generate an embedding representation $H_{T}\in \mathbb{R} ^{ N\times T\times D_{H}}$ to capture temporal dependencies. The computation process is as follows:
\begin{align}
    &\tilde{H}_{Trend}=FC(H_{Trend}),\\
    &\tilde{H}_{Remainder}=FC(H_{Remainder}),\\
    &H_{T}= \tilde{H}_{Trend}\oplus \tilde{H}_{Remainder}.
\end{align}%
\par Apart from temporal dependencies, the spread of epidemics is also affected by spatial information among patches. To incorporate this spatial impact,  we utilize the temporal embedding $H_{T}$ and the dynamic graph $A_{dyn}$ as inputs to the GCN model. This enables us to generate an embedding representation $H_{ST}\in \mathbb{R} ^{ N\times T\times D_{ST}}$ that effectively captures both temporal and spatial dependencies, where $D_{ST}$ denotes the dimension of GCN:
\begin{align}
    H_{ST}&=GCN(H_{T},A_{dyn}),\nonumber \\
    &=A_{dyn}H_{T}W,
\end{align}%
where $W\in \mathbb{R} ^{ N\times T\times D_{ST}}$ denotes the learnable parameters, and $A_{dyn}$ has been normalized and can be used directly for GCN aggregation of neighbor information.
\subsection{Epidemiology Module}
\par In epidemic forecasting,~\cite{adiga2022ai} points out that using only spatio-temporal models may not be sufficient for accurate forecasting because these models lack an understanding of the underlying physical phenomena in epidemic evolution. Given the abundance of domain knowledge models in the field of epidemiology, researchers often incorporate these domain models into neural networks to assist in learning the potential dynamics of epidemic spread~\cite{gao2021stan,wang2022causalgnn,cao2023mepognn,mao2023mpstan}. This integration provides more accurate epidemic forecasting and deeper insights.
\par Therefore, in the epidemiology module, we introduce the SIR model as epidemiology domain knowledge and integrate it into the neural network to achieve more accurate forecasting. The SIR model divides the population into three labeled compartments: susceptible (S), infected (I), and recovered (R), and represents the interaction between these compartments through the following three differential equations~\cite{sattenspiel1995structured}:
\begin{align}
    &\frac{dS_{i}}{dt}=-\beta_{i}I_{i} \frac{S_{i} }{N_{i} },\\
    &\frac{dI_{i} }{dt}=\beta_{i}I_{i}\frac{S_{i} }{N_{i} }-\gamma_{i}I_{i}, \\
    &\frac{dR_{i} }{dt}=\gamma_{i} I_{i},
\end{align}%
where $S_{i}$, $I_{i}$, $R_{i}$, and $N_{i}$ represent the number of susceptible, infected, recovered individuals, and the total population in patch $i$, respectively. Additionally, $\beta_{i}$ and $\gamma_{i}$ represent the infection rate and recovery rate of patch $i$, respectively.
\par To incorporate the SIR model into the neural network, we feed the embedding representation $H_{ST}$, which captures spatio-temporal dependencies, into a fully connected layer. This layer is designed to learn the infection and recovery rates $\beta, \gamma \in \mathbb{R} ^{ N}$ that govern the dynamics of epidemic transmission and normalize them by using the Sigmoid function:
\begin{align}
    &\beta,\gamma =FC(Reshape(H_{ST})),\\
    &\beta=Sigmoid(\beta), \\
    &\gamma=Sigmoid(\gamma),
\end{align}%
where $Reshape()$ is used to reshape the dimensions of embeddings.
\par Finally, by integrating the original epidemic data $X_{T}$ at time step $T$, infection rate $\beta$, and recovery rate $\gamma$ with the SIR model, we can utilize domain knowledge models to infer the future changes in the epidemic dynamics for each patch:
\begin{align}
    \Delta X_{Phy,T}  =SIR(X_{T},\beta,\gamma),
\end{align}%
where $\Delta X_{phy, T}\in \mathbb{R} ^{ N\times 3}$ represents the changes in the number of susceptible, infected, and recovered individuals across all patches at time $T$.

\begin{table*}
    \centering
    \begin{tabular}{llllllll}
        \toprule
        Dataset& Data level & Data size & Time range & Min &Max   &  Mean & Std \\
        \midrule
        US & State-level & $52\times245$ & 2020.5.1-2020.12.31& 0 & 838855& 40438& 75691  \\
        Japan &Prefecture-level & $47\times151$ & 2022.1.15-2022.6.14 & 104 & 198011 & 11458 & 21188 \\
        \bottomrule
    \end{tabular}
    \caption{Statistical information of the datasets.}
    \label{table:data}
\end{table*}
\subsection{Forecasting Module}
\par The output of BDGSTN is divided into two categories: neural network output and epidemiological output. The neural network output is generated by using the embedding representation $H_{ST}$, which captures spatio-temporal dependencies, as input to a fully connected layer. This predicts the number of infected individuals $\widehat{Y}^{ST}\in\mathbb{R} ^{ N\times L}$ for all patches in the future $L$ time steps: 
\begin{align}
    \widehat{Y}^{ST}=FC(Reshape(H_{ST})).
\end{align}%
\par The epidemiological output is generated by iteratively computing $\Delta X$ using the SIR model to predict the number of infected individuals for all patches in the future $L$ time steps. The calculation is as follows: 
\begin{align}
    &X_{Phy,T+1}=X_{T}+\Delta X_{Phy,T},\\
    &X_{Phy,T+1}=[X^{S}_{Phy,T+1},X^{I}_{Phy,T+1},X^{R}_{Phy,T+1}],\\
    &\dots \nonumber \\
    &\widehat{Y}^{Phy}=[X^I_{Phy,T+1},X^I_{Phy,T+2},\dots,X^I_{Phy,T+L}],
\end{align}%
where $X_{Phy,T+1}\in \mathbb{R} ^{ N\times 3}$ denotes the epidemiological forecasting for each state at time step $T+1$, and $\widehat{Y}^{Phy}\in \mathbb{R} ^{ N\times L}$ represents the number of infected individuals in the future $L$ time steps based on the SIR model. 
\subsection{Optimization}
\par We choose Mean Absolute Error (MAE) as the objective function and simultaneously compare the forecasted $\widehat{Y}^{ST}$ and $\widehat{Y}^{Phy}$ of the neural network and epidemiological model with the ground truth $Y$ to jointly constrain the model training. The objective function to be minimized is formulated as follows:
\begin{align}
    \mathcal{L} (\Theta )=\frac{1}{N\times L }  \sum_{i=1}^{N}\sum_{\tau =1}^{L}(\left |\widehat{Y}^{ST}_{i,\tau }-{Y}_{i,\tau }\right |+ \left |\widehat{Y}^{Phy}_{i,\tau }-{Y}_{i,\tau } \right | ).
\end{align}%

\section{Experiments} \label{sec:Experiments}

\subsection{Datasets}
\par Our experiments are conducted on two real datasets: the US dataset and the Japan dataset. These datasets include the number of daily infected cases, daily recovered cases, and daily susceptible cases over some time. Daily infected cases are used as the primary feature, while the others serve as auxiliary features. The statistical information of the datasets is presented in \reftable{table:data}, and other details are provided below:
\begin{enumerate}[(1)]
  \item \textbf{US}: This dataset is a state-level dataset collected from the Johns Hopkins Coronavirus Resource Center~\cite{dong2020interactive} that records population numbers and Covid-related data for each state from May 1, 2020 to December 31, 2020 (245 days).
  \item \textbf{Japan}: This dataset is a prefecture-level dataset collected from the Japan LIVE Dashboard~\cite{su2021japan} that records population numbers and Covid-related data for each state from January 15, 2022 to June 14, 2022 (151 days).
\end{enumerate}

\begin{table*}
    \centering
    \resizebox{\textwidth}{!}{
    \begin{tabular}{lllllllllll}
      \toprule
        &\multicolumn{10}{l}{The US dataset}\\
        \cmidrule(l){2-11}
        &\multicolumn{5}{l}{$L$=5}&\multicolumn{5}{l}{$L$=10}\\

        \cmidrule(l){2-6}\cmidrule(l){7-11}
         Model & MAE & RMSE &MAPE & PCC & CCC  & MAE & RMSE &MAPE & PCC & CCC  \\
        \midrule 
        SIR  &5660  & 15656 &25.81\%  & 99.16\%&  99.14\%  & 10608 &  33766& 41.94\% &  96.23\% & 96.09\%  \\
        ARIMA &6475  &22095  & 14.01\% &98.33\%  &98.31\%  &11489  & 44779 &26.36\%  & 93.66\% & 93.39\% \\
        GRU & 18348 &32950  & 21.88\% & 97.88\% & 95.63\% & 26749 & 47328 &32.52\%  &95.66\%  &90.39\%   \\
        GraphWaveNet & 13875 &22559  & 17.85\% & 99.46\% & 97.82\% & 9526 & 15673 & 16.64\% &99.21\% & 99.09\% \\
        STGODE & 70454 &116865  &  83.21\%& 91.95\% & 64.48\% &53693  &83823  & 63.51\% & 87.89\% &  62.19\%\\
        CovidGNN & 9453 & 21612 &9.91\%  & 99.07\% &  98.17\%& 16052 &37586  & 15.03\% & 96.87\% & 94.00\% \\
        ColaGNN & 66005 &  111622&   77.57\% &  53.54\%& 41.79\% & 51822 & 91680 & 57.61\% & 80.18\%&62.46\%  \\
        STAN  &10024  & 19214 & 17.98\% &98.70\% & 98.65\% & 13993 &25963  &  19.38\%& 97.80\% & 97.49\% \\
        MPSTAN & \underline{3960} &\underline{8255} & \underline{6.38\%} & \underline{99.80\%}& \underline{99.75\%} &\underline{7711}  & \underline{14463} & \underline{10.73\%}& \underline{99.55\%} & \underline{99.20\%} \\
        BDGSTN & \textbf{3196} & \textbf{7471}& \textbf{6.09\%} & \textbf{99.81\%}& \textbf{99.80\%} & \textbf{5619} & \textbf{11751} & \textbf{9.23\%}& \textbf{99.56\%} & \textbf{99.50\%} \\
        Improvement &19.97\%  & 9.50\% &4.55\%  & 0.01\% &0.05\%  & 27.13\% & 18.75\% & 13.98\%& 0.01\% &  3.02\%\\
        \toprule
        &\multicolumn{5}{l}{$L$=15}&\multicolumn{5}{l}{$L$=20}\\

        \cmidrule(l){2-6}\cmidrule(l){7-11}
          Model& MAE & RMSE &MAPE & PCC & CCC  & MAE & RMSE &MAPE & PCC & CCC  \\
        \midrule 
        SIR  &16573  & 60984 &57.38\%  &89.04\%  &88.26\%  &23963  &101612  &76.12\%  & 76.44\% & 73.21\% \\
        ARIMA &  17151& 74295 & 43.01\% &84.86\%  & 83.59\% & 24849 &121875  & 65.20\% & 69.78\% & 65.31\% \\
        GRU & 33968 & 59804 & 41.21\% & 92.67\% & 83.94\% & 38202 & 65762 & 45.54\% & 90.44\% &80.61\%   \\
        GraphWaveNet &47020  &76735  & 51.64\% & 90.76\% & 72.64\% & 48154 &82098  & 51.19\% & 84.43\% & 68.47\% \\
        STGODE &72622  &117611  & 107.65\% & 82.26\% & 50.75\% &72132  & 109536 & 84.84\% & 85.16\% & 42.81\%  \\
        CovidGNN & 21660 &48169 &19.85\% & 94.68\% & 89.71\% &26985  &57085  & 24.57\% & 92.64\% & 84.95\%  \\
        ColaGNN &33419  & 55424 & 41.36\% &92.43\%  &79.63\%  &47837  & 77656 & 52.48\% & 92.49\% & 70.90\% \\
        STAN  & 16784 &33383  &20.78\%  &96.43\%  &  95.49\%& 18679 & 36180 & 26.81\% & 96.09\% &  94.52\% \\
        MPSTAN &  \underline{10148}&  \underline{18460}& \underline{14.68\%} & \textbf{99.25\%} &  \underline{98.68\%}  & \underline{12728} & \underline{22923} & \underline{18.68\%} & \textbf{98.81\% }& \underline{97.91\%}\\
        BDGSTN & \textbf{8253} & \textbf{16193}& \textbf{13.57\%} & \underline{99.22\%}& \textbf{99.01\%} & \textbf{11139} & \textbf{21304} & \textbf{18.06\%}& 98.78\% & \textbf{98.23\%} \\
        Improvement &18.67\%  & 12.28\% &7.56\%  & - &0.33\%  & 12.48\% & 7.06\% & 3.32\%& - &  0.33\%\\
          \bottomrule
    \end{tabular}
    }
    \caption{Performance comparison with baseline models on the US dataset.}
    \label{table:us-exp}
\end{table*}

\subsection{Baseline Models}
\par We compare BDGSTN with the following five types of algorithms: (\romannumeral1) traditional mathematical methods: SIR, ARIMA, (\romannumeral2) time-series models: GRU, (\romannumeral3) traffic spatio-temporal models: GraphWaveNet, STGODE, (\romannumeral4) traditional epidemic spatio-temporal models: CovidGNN, ColaGNN, and (\romannumeral5) epidemic spatio-temporal models that incorporate domain knowledge: STAN, MPSTAN.
\begin{enumerate}[(1)]
  \item \textbf{SIR}~\cite{sattenspiel1995structured}: The SIR model utilizes three differential equations in conjunction with real data to simulate the future changes in the number of individuals in different states of an epidemic.
  \item \textbf{ARIMA}~\cite{hernandez2020forecasting}: The Auto-Regressive Integrated Moving Average model is a well-known statistical model for time-series analysis.
  \item \textbf{GRU}~\cite{chung2014empirical}: The GRU model uses few parameters and introduces gate mechanisms in its recurrent structure to efficiently control the generation of time-series data.
  \item \textbf{GraphWaveNet}~\cite{wu2019graph}: GraphWaveNet model employs learnable embeddings to construct a graph structure and combines diffusion GCN and gated TCN to capture the spatio-temporal dependencies.
  \item \textbf{STGODE}~\cite{fang2021spatial}: STGODE proposes a continuous representation of GCN based on NeuralODE to extract long-term spatio-temporal correlations.
  \item \textbf{CovidGNN}~\cite{kapoor2020examining}: CovidGNN takes the time-series data of each patch as features and utilizes a GCN with skip connections to predict future numbers of infected individuals.
  \item \textbf{ColaGNN}~\cite{deng2020cola}: ColaGNN integrates a dynamic position-aware attention mechanism and temporal dilation convolution to jointly predict real influenza data.
  \item \textbf{STAN}~\cite{gao2021stan}: STAN utilizes the SIR model to construct a dynamic constraint loss in spatio-temporal models, thereby assisting in the training process of the model.
  \item  \textbf{MPSTAN}~\cite{mao2023mpstan}: MPSTAN introduces a metapopulation epidemic model called MP-SIR and incorporates it into the deep learning model construction and loss functions to enhance the learning of the underlying dynamics.
\end{enumerate}

\begin{table*}
    \centering
    \resizebox{\textwidth}{!}{
    \begin{tabular}{lllllllllll}
      \toprule
        &\multicolumn{10}{l}{The Japan Dataset}\\
        \cmidrule(l){2-11}
        &\multicolumn{5}{l}{$L$=5}&\multicolumn{5}{l}{$L$=10}\\

        \cmidrule(l){2-6}\cmidrule(l){7-11}
          Model& MAE & RMSE &MAPE & PCC & CCC  & MAE & RMSE &MAPE & PCC & CCC  \\
        \midrule 
        SIR  & \textbf{896} &  \textbf{1572}& 18.89\% & \textbf{99.11\%} &  \textbf{97.91\%}&  1703& \textbf{2874} & 39.38\% &\textbf{97.73\%}  & \textbf{93.67\%}  \\
        ARIMA & 1113 & 3137 & 24.33\% & 91.74\% & 91.37\% & 2433 & 8719 & 59.59\% & 63.42\% & 57.19\%  \\
        GRU   & 2156 & 3955 &58.91\%  &94.06\%  &89.02\%  & 2702 & 5130 & 69.49\% & 92.33\% & 83.80\% \\
        GraphWaveNet  & 2048 &4490  & 39.06\% & 94.93\% &87.35\%  & 2744 & 6447 &  48.88\%& 92.64\% & 79.24\% \\
        STGODE  &  5420& 13057 &103.14\%  & 83.94\% &57.16\%  &  8208&  18396& 158.08\% &85.00\%  &50.91\%   \\
        CovidGNN  &1042  & 2305 & 18.06\% & 97.27\% & 95.71\% &1887  & 3942 &39.40\%  &95.77\%  & 89.48\% \\
        ColaGNN   & 2566 & 5746 &50.29\%  & 92.17\% &82.16\%  & 5294 &  10402& 101.50\% &86.60\%  & 63.78\% \\
        STAN &1070  &  2400&22.97\%  & 95.87\% &94.82\%  &1623  & 3165 & 34.38\% & 94.80\%& 91.97\%  \\
        MPSTAN  &  1016& 2311 & \textbf{16.91\%} &96.74\%  &95.60\%  &\textbf{1356}  &  \underline{3016}& \textbf{24.34\%}&93.38\%  & 92.27\% \\
        BDGSTN & \underline{922} & \underline{1922}& \underline{18.07\%} & \underline{98.34\%}& \underline{96.93\%} &\underline{1407}  & 3238 & \underline{25.74\%}& \underline{96.27\%} & \underline{92.32\%} \\
        Improvement &- & -&- & - &- & -& -& -& - &  -\\
        \toprule
        \multirow{4}*{Model}&\multicolumn{5}{l}{$L$=15}&\multicolumn{5}{l}{$L$=20}\\

        \cmidrule(l){2-6}\cmidrule(l){7-11}
          & MAE & RMSE &MAPE & PCC & CCC  & MAE & RMSE &MAPE & PCC & CCC  \\
        \midrule 
        SIR  & 2632 & 4373 &66.60\%  & \underline{95.22\%} & 87.05\% & 3515 & 5883 &92.93\%  &92.08\%  &   79.20\% \\
        ARIMA & 3443 & 7715 & 86.16\% & 65.62\% & 61.39\% &3757  & 7513 &130.90\%  &72.79\%  & 66.56\% \\
        GRU   &  2124&3758  &59.84\%  & 88.58\% &  87.70\%& 2977 &5343  &68.13\%  &  71.75\%&  70.72\%  \\
        GraphWaveNet &2828  &  6520&  49.39\%& 93.62\% & 79.34\% & 2773 & 6547  & 46.11\% &  \textbf{92.96\%}& 79.38\% \\
        STGODE &10330  &23345  &195.76\% & 82.22\% & 38.62\% & 12156 & 27407 & 221.51\% & 83.58\% & 33.33\%   \\
        CovidGNN  &2988  &  6515& 66.73\% & 90.20\% & 77.42\% &3990  &  8805& 94.82\% &84.97\%  & 67.12\% \\
        ColaGNN    &4192  & 8688 & 93.21\% &84.31\%  & 67.68\% & 7195 &15400  & 140.32\% & 84.30\% & 50.40\%  \\
        STAN & 2026 & 3887 & 51.03\% & 93.86\% & 88.92\% & 2804 & 5238 & 72.10\% & 90.59\%& 82.24\%  \\
        MPSTAN &  \underline{1465}& \underline{3104} &\underline{28.29\% } & 91.84\% & \underline{91.29\%} & \underline{1854} & \underline{4014} &\underline{34.67\%}  &  85.78\%& \underline{84.97\%}  \\
        BDGSTN & \textbf{1272} &\textbf{2833} & \textbf{26.97\%} & \textbf{95.35\%}& \textbf{93.40\%} &  \textbf{1313}& \textbf{2897} & \textbf{28.24\%}& \underline{92.48\%} & \textbf{92.28\%} \\
        Improvement &13.17\%  & 8.73\% &4.67\%  & 0.14\% &2.31\%  & 29.18\% & 27.83\% & 18.55\%& - &  8.60\%\\
          \bottomrule
    \end{tabular}
    }
    \caption{Performance comparison with baseline models on the Japan dataset.}
    \label{table:japan-exp}
\end{table*}

\begin{figure*}[t]
	\centering  
	\subfigure[Forecasting curve for US-Indiana]{
		\includegraphics[width=0.48\linewidth]{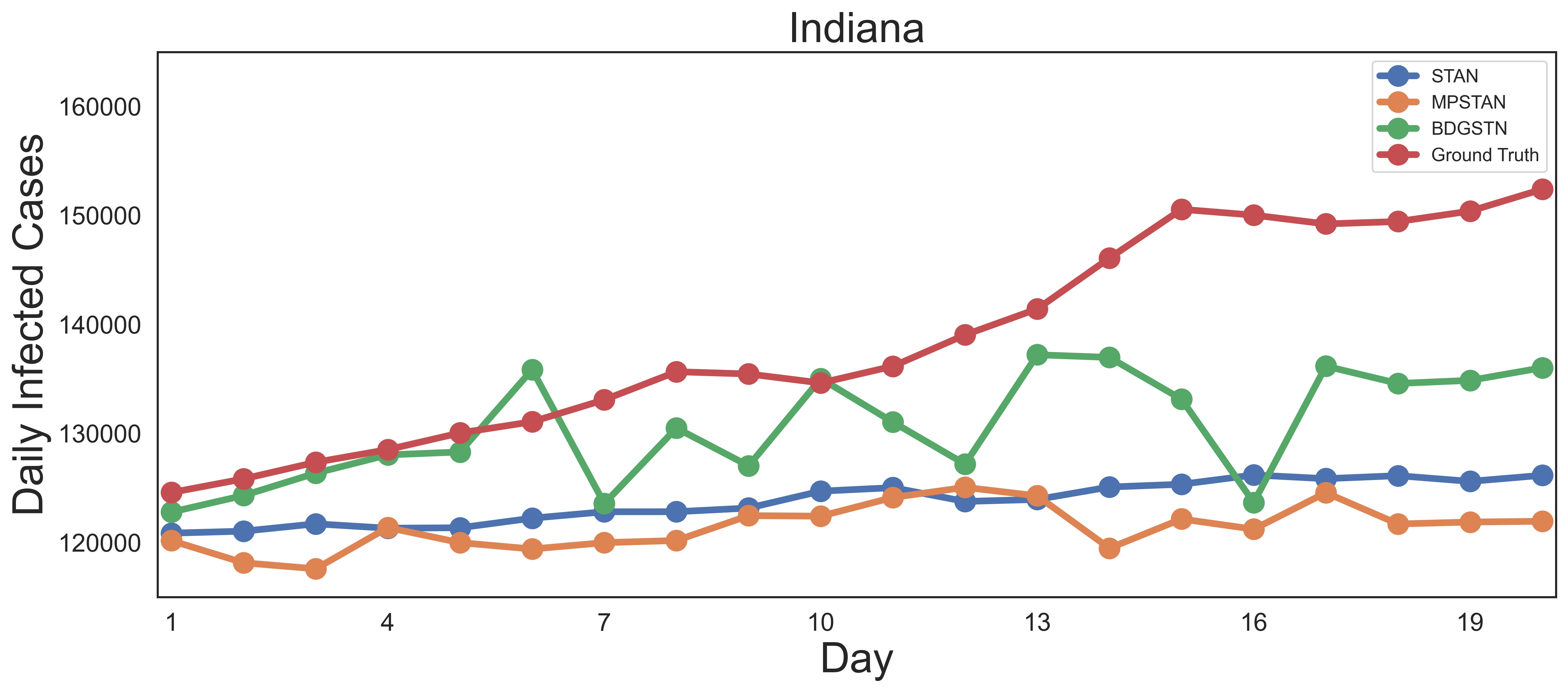}}
        \subfigure[Forecasting curve for US-Maryland]{
		\includegraphics[width=0.48\linewidth]{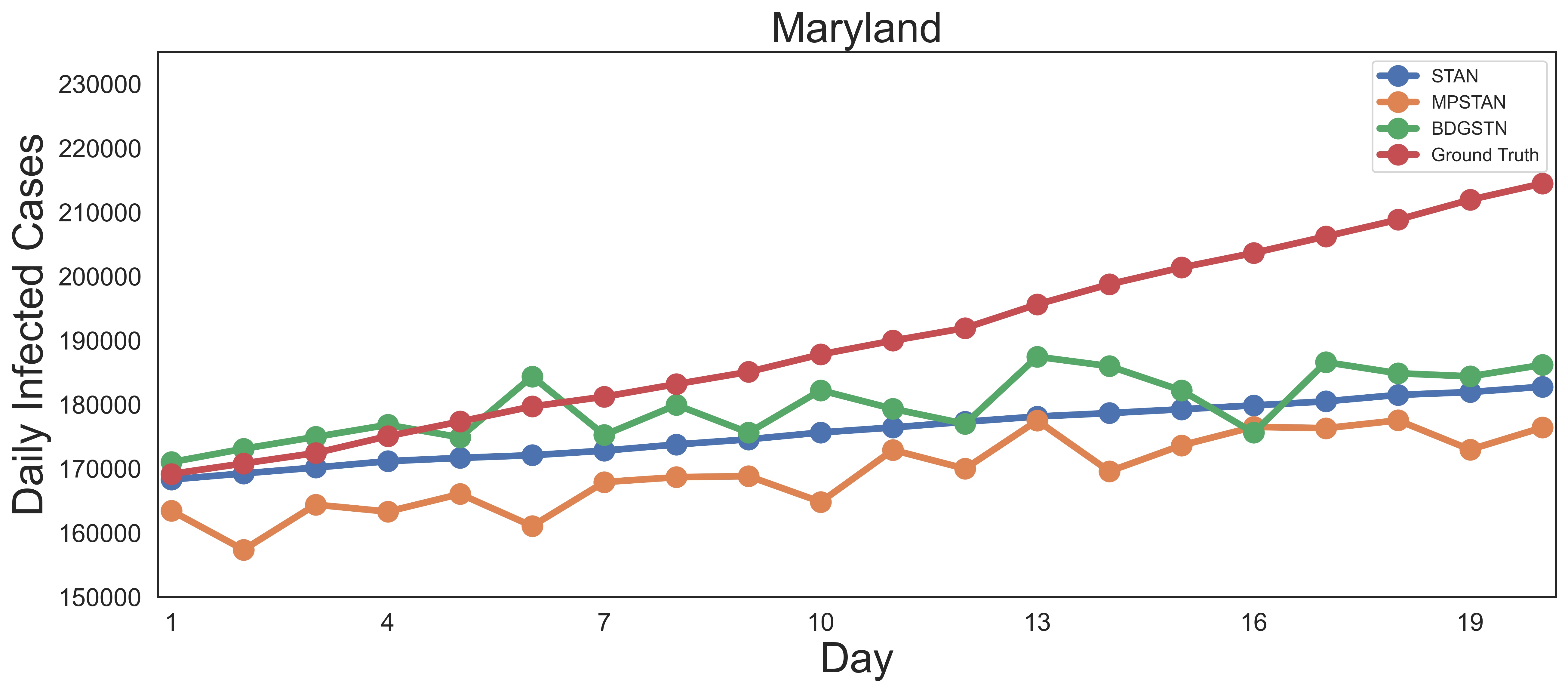}}
        \subfigure[Forecasting curve for Japan-Toyama]{
		\includegraphics[width=0.48\linewidth]{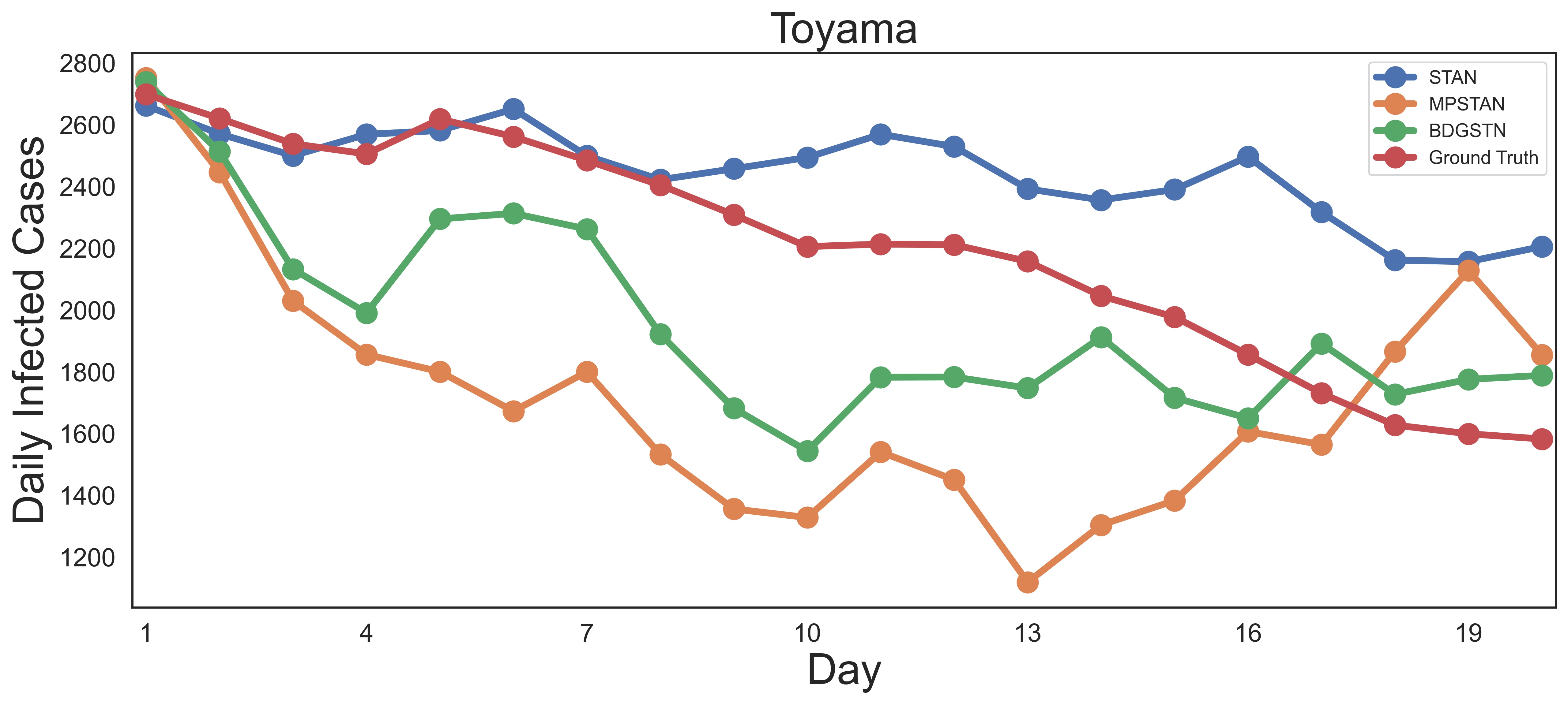}}
        \subfigure[Forecasting curve for Japan-Gifu]{
		\includegraphics[width=0.48\linewidth]{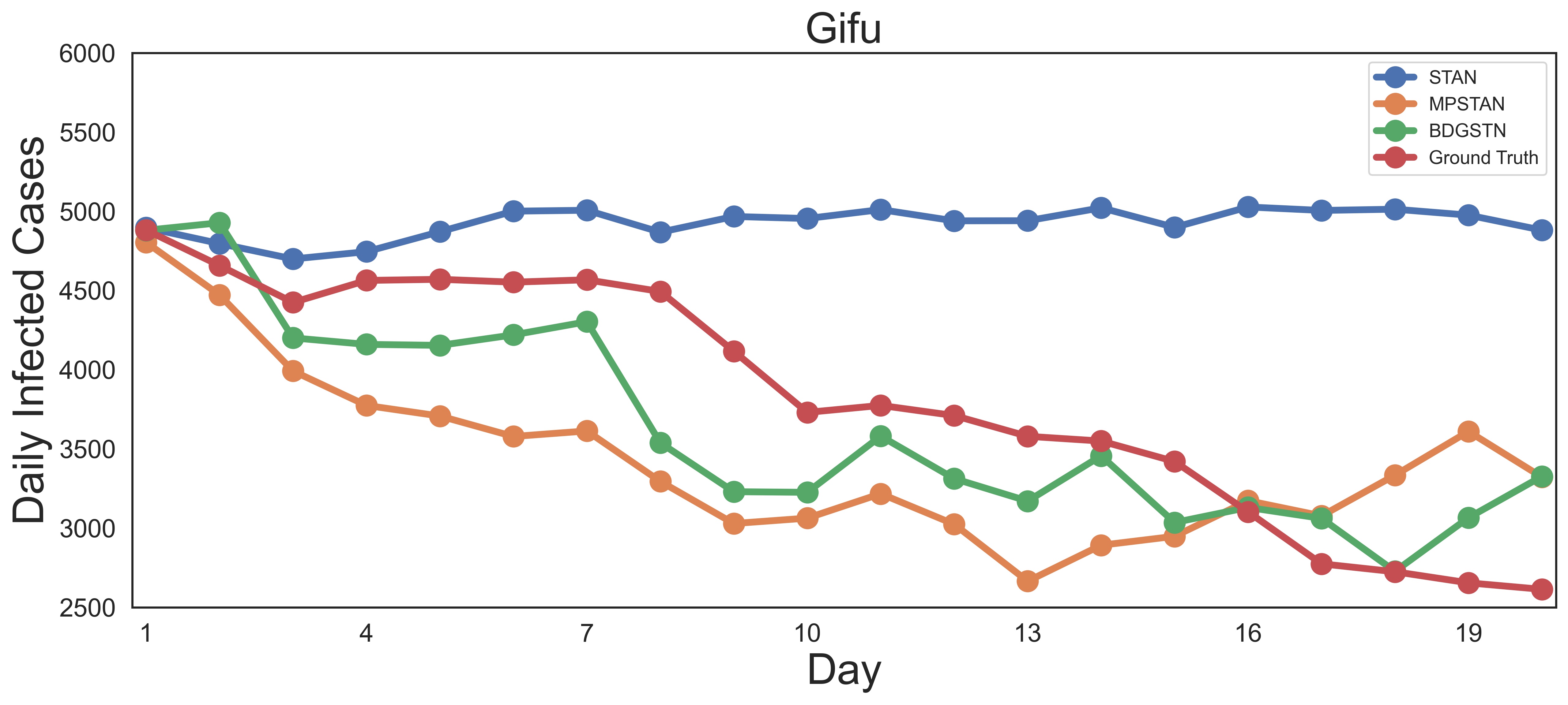}}
	\caption{Visualization of forecasting curves in US(a)(b) and Japan(c)(d).}
    \label{fig3}
\end{figure*}

\subsection{Setup of Experiments}
\par We divide each of the two datasets into training, validation, and test sets in a ratio of 60\%-20\%-20\% and normalize all the data to the range of (0, 1). To validate the effectiveness of BDGSTN in short-term and long-term forecasting, we set the input time length to 5 and set the output time length to 5 and 10 for short-term forecasting, and 15 and 20 for long-term forecasting. In BDGSTN, we set the initial embedding dimension $D_{H}$, DLinear, and GCN dimensions to 32. For dynamic graph learning, the embedding dimension $D_{ada}$ of the backbone graph is also set to 32. In the temporal graph, we utilize a 1-layer TCN model with an embedding dimension $D_{TCN}$ of 8 and a kernel size of 3. Additionally, we set epoch numbers at 200 and use the Adam optimizer with a learning rate of 1e-4.
\par To evaluate the performance of the model, we employ several evaluation metrics, namely MAE, RMSE, MAPE, PCC, and CCC. The lower the MAE, RMSE, and MAPE, and the higher the PCC and CCC, the better the performance. The above evaluation metrics are expressed as follows:
\begin{align}
    &MAE=\frac{1}{N\times L} \sum_{i=1}^{N}\sum_{\tau=1}^{L}  |\widehat{Y}_{i,\tau}^{ST} -Y_{i,\tau } |,\\
    &RMSE=\sqrt{\frac{1}{N \times L} \sum_{i=1}^{N}\sum_{\tau=1}^{L}|\widehat{Y}_{i,\tau}^{ST} -Y_{i,\tau } |^{2} }, \\
    &MAPE=\frac{100\%}{N \times L} \sum_{i=1}^{N}\sum_{\tau=1}^{L}\left | \frac{\widehat{Y}_{i,\tau}^{ST} -{Y}_{i,\tau }}{Y_{i,\tau }} \right |  , \\
    &PCC=\frac{\sum(\widehat{Y}_{i,\tau}^{ST}-\bar{ \widehat{Y}}_{i,\tau}^{ST} )({Y}_{i,\tau}
-\bar{{Y}}_{i,\tau })}{\sqrt{\sum (\widehat{Y}_{i,\tau}^{ST}-\bar{ \widehat{Y}}_{i,\tau}^{ST} )^{2}\sum (Y_{i,\tau}
-\bar{Y}_{i,\tau })^{2}} },\\
    &CCC=\frac{2\rho \sigma _{x}\sigma _{y}}{\sigma _{x}^{2}+\sigma _{y}^{2}+(\mu _{x}-\mu _{y})^{2}} ,
\end{align}
where $\rho$ denotes the correlation coefficient between the two variables, $\mu _{x}$ and $\mu _{y}$ denote the mean of the two variables, $\sigma _{x}^{2}$ and $\sigma _{y}^{2}$ are the corresponding variances, and $\sum$ is an abbreviation for $\sum_{i=1}^{N}\sum_{\tau=1}^{L}$.
\subsection{Forecasting Performance}

\par In this experiment, we compare the BDGSTN model with nine baseline models for short-term and long-term epidemic forecasting. The comparison results are shown in \reftable{table:us-exp} and \reftable{table:japan-exp}, where bold indicates the model achieves State-of-the-Art (SOTA) forecasting performance, underlining indicates suboptimal forecasting performance, and Improvement represents the improvement of BDGSTN compared to the suboptimal performance. According to the results, the BDGSTN model mostly achieves SOTA or the most competitive results in different forecasting tasks on various datasets compared to the other models.
\par On the US dataset, the BDGSTN model achieves SOTA performance and exhibits significant improvement compared to the suboptimal model. Specifically, the BDGSTN model achieved at least 12.48\% improvement in MAE, 7.06\% improvement in RMSE, 3.32\% improvement in MAPE, and 0.05\% improvement in CCC, as shown in \reftable{table:us-exp}. \reftable{table:japan-exp} presents the predictive performance on the Japan dataset. We can observe that the BDGSTN model consistently delivers stable suboptimal performance in short-term forecasting ($L$=5, 10), with only a small performance gap compared to the SOTA model, making its short-term forecasting highly competitive. In long-term forecasting ($L$=15, 20), the BDGSTN model achieved SOTA performance in all tasks, with at least 13.17\% improvement in MAE, 8.73\% improvement in RMSE, 4.67\% improvement in MAPE, and 2.31\% improvement in CCC. Overall, the BDGSTN model consistently provides SOTA or competitive predictive results compared to the other models.

\par In addition, we select two locations from the datasets in the US and Japan, and utilize high-performing models to visualize the comparison between the forecasted and actual values when $L$=20. \reffig{fig3}(a)(b) represent the forecasting curves for Indiana and Maryland in the US, respectively, while \reffig{fig3}(c)(d) represent the forecasting curves for Toyama and Gifu in Japan, respectively. It can be observed that the forecasted values of BDGSTN are fairly close to the actual values, as evident in \reffig{fig3}. However, as the forecasting window increases, the forecasting accuracy decreases, as shown in \reffig{fig3}(a)(b). This is also consistent with the data in \reftable{table:us-exp} and \reftable{table:japan-exp}. As the forecasting window increases, the model faces challenges in accurate long-term forecasting. Nevertheless, BDGSTN exhibits superior performance in long-term forecasting compared to other models. 

\par Next, we compare the performance between different models. Firstly, it is worth noting that traditional mathematical models (such as SIR and ARIMA) may outperform deep learning models in early epidemic forecasting, but their performance declines in long-term forecasting. Specifically, in the Japan dataset, the SIR model demonstrates SOTA performance in short-term forecasting, better than BDGSTN. This may be because the SIR model is used to calculate the change in the number of infected individuals at each time step, which is beneficial for short-term forecasting. However, as the forecasting window becomes longer, errors gradually accumulate, leading to a decline in long-term forecasting performance.
\par The GRU model is unable to provide more accurate epidemic prediction than the other epidemic spatio-temporal models, possibly because it does not consider the spatial impact of epidemic transmission. In addition, traffic spatio-temporal models (such as GraphWaveNet and STGODE) struggle to provide stable and accurate forecasting when applied to epidemics. This could be attributed to the fact that epidemic data is sparser and noisier compared to traffic data, increasing the possibility of overfitting for these models. 
\par Further discussing the spatio-temporal epidemic model, ColaGNN may not be suitable for larger-scale and more complex COVID-19 data, as described in \cite{mao2023mpstan}, since it was originally designed for influenza-like illnesses. In addition, spatio-temporal models that incorporate domain knowledge (such as STAN, MPSTAN, and BDGSTN) perform better than the other types of models, with BDGSTN having the best performance. The superior performance of BDGSTN can be attributed not only to the constraints of domain knowledge but also to the effectiveness of dynamic graph learning. The effectiveness of dynamic graph learning will be discussed in sections \ref{sub:4.7} and \ref{sub:4.8}.

\subsection{Effects of Spatio-Temporal Dependencies}
\par To explore the impact of temporal and spatial dependencies on spatio-temporal epidemic forecasting, we design two variant models: BDGSTN-Temporal and BDGSTN-Spatial.

\begin{table*}
    \centering
    \resizebox{\textwidth}{!}{
    \begin{tabular}{lllllllllll}
      \toprule
        &\multicolumn{10}{l}{The US dataset}\\
        \cmidrule(l){2-11}
        &\multicolumn{5}{l}{$L$=5}&\multicolumn{5}{l}{$L$=10}\\

        \cmidrule(l){2-6}\cmidrule(l){7-11}
         Model & MAE & RMSE &MAPE & PCC & CCC  & MAE & RMSE &MAPE & PCC & CCC  \\
        \midrule 
        BDGSTN-Temporal &3432  &7799  & 6.23\% & 99.80\% & 99.78\%& 6302 & 12586 & 9.97\% &99.54\%  & 99.42\% \\
        BDGSTN-Spatial&4198  & 9285 & 7.10\% & 99.70\% &99.70\% & 6245 &12733  &10.02\%  & 99.48\% & 99.41\% \\
        BDGSTN & \textbf{3196} & \textbf{7471}& \textbf{6.09\%} & \textbf{99.81\%}& \textbf{99.80\%} & \textbf{5619} & \textbf{11751} & \textbf{9.23\%}& \textbf{99.56\%} & \textbf{99.50\%} \\
        \toprule
        &\multicolumn{5}{l}{$L$=15}&\multicolumn{5}{l}{$L$=20}\\

        \cmidrule(l){2-6}\cmidrule(l){7-11}
          Model& MAE & RMSE &MAPE & PCC & CCC  & MAE & RMSE &MAPE & PCC & CCC  \\
        \midrule 
        BDGSTN-Temporal & 9369 &17750  & 14.33\% &99.19\%  &98.79\% &  12460&  23134& 19.08\% & 98.76\% & 97.87\% \\
        BDGSTN-Spatial& 8581 & 16694 & 14.31\% & 99.17\% & 98.95\%& 11470 & 21966 & 19.04\% &98.67\%  &  98.12\%\\
        BDGSTN & \textbf{8253} & \textbf{16193}& \textbf{13.57\%} & \textbf{99.22\%}& \textbf{99.01\%} & \textbf{11139} & \textbf{21304} & \textbf{18.06\%}& \textbf{98.78\%} & \textbf{98.23\%} \\
          \bottomrule
    \end{tabular}
    }
    \caption{Effects of spatio-temporal dependencies on the US dataset.}
    \label{table4}
\end{table*}
\begin{table*}
    \centering
    \resizebox{\textwidth}{!}{
    \begin{tabular}{lllllllllll}
      \toprule
        &\multicolumn{10}{l}{The Japan Dataset}\\
        \cmidrule(l){2-11}
        &\multicolumn{5}{l}{$L$=5}&\multicolumn{5}{l}{$L$=10}\\

        \cmidrule(l){2-6}\cmidrule(l){7-11}
         Model & MAE & RMSE &MAPE & PCC & CCC  & MAE & RMSE &MAPE & PCC & CCC  \\
        \midrule 
        BDGSTN-Temporal & \textbf{787} & \textbf{1705} & \textbf{13.97\%} & \textbf{98.68\%} &\textbf{97.58\%} & \textbf{1026} & \textbf{2246} & \textbf{19.13\%} & \textbf{97.72\%} & \textbf{95.89\%} \\
        BDGSTN-Spatial& 1222 & 2766 &22.59\%  & 97.12\% &94.13\% & \textbf{1278} &\textbf{2970}  &\textbf{24.14\%}  & 96.12\% &  \textbf{93.21\%}\\
        BDGSTN & 922 & 1922& 18.07\% & 98.34\%& 96.93\% &1407  & 3238 & 25.74\%& 96.27\% & 92.32\% \\
        \toprule
        &\multicolumn{5}{l}{$L$=15}&\multicolumn{5}{l}{$L$=20}\\

        \cmidrule(l){2-6}\cmidrule(l){7-11}
          Model& MAE & RMSE &MAPE & PCC & CCC  & MAE & RMSE &MAPE & PCC & CCC  \\
        \midrule 
        BDGSTN-Temporal &\textbf{1069}  & \textbf{2161} &\textbf{20.65\%}  & \textbf{95.54\%} & \textbf{94.89\%}& 1747 & 3592 & 29.73\% &87.93\%  & 84.89\% \\
        BDGSTN-Spatial& 1448 & 3396 &30.11\%  & 93.14\% & 90.80\%& 1470 &  3316& 31.12\% &90.30\%  &90.00\%  \\
        BDGSTN & 1272 &2833 & 26.97\% & 95.35\%& 93.40\% &  \textbf{1313}& \textbf{2897} & \textbf{28.24\%}& \textbf{92.48\%} & \textbf{92.28\%} \\
          \bottomrule
    \end{tabular}
    }
    \caption{Effects of spatio-temporal dependencies on the Japan dataset.}
    \label{table5}
\end{table*}

\begin{enumerate}[(1)]
  \item \textbf{BDGSTN-Temporal}: BDGSTN-Temporal model removes the spatial information processing part from BDGSTN, including dynamic graph learning and GCN, and focuses solely on handling the temporal information of epidemics.
  \item \textbf{BDGSTN-Spatial}: BDGSTN-Spatial model disregards the temporal information in BDGSTN, removes the DLinear model, and is solely used for aggregating the spatial impact of epidemics.
\end{enumerate}
\par The experimental results on the datasets from the US and Japan are shown in \reftable{table4} and \reftable{table5}. According to the records in \reftable{table4}, BDGSTN demonstrates the best performance on the US dataset by adequately considering the temporal and spatial dimensions, which provides better assistance in epidemic forecasting. However, on the Japan dataset, as indicated in \reftable{table5}, we observe that the BDGSTN-Temporal model, which considers only the temporal dimension, achieves better performance for tasks with $L$=5, 10, and 15. This finding aligns with the analysis conducted by \cite{mao2023mpstan}, suggesting that the Japan dataset exhibits limited population mobility, leading to a diminished influence of spatial factors. Consequently, focusing solely on the temporal dimension may yield superior results. However, relying solely on the temporal dimension may lead to less stable outcomes. For instance, at $L$=20, we observe a significant performance improvement of BDGSTN compared to the BDGSTN-Temporal model. Overall, compared to the BDGSTN-Temporal model, BDGSTN provides stable and competitive forecasting results. Compared to the BDGSTN-Spatial model, BDGSTN consistently performs better across different datasets. This is because epidemic data fundamentally exhibits temporal characteristics, and disregarding the temporal dependencies inevitably leads to a decline in forecasting accuracy. Therefore,  it is crucial to consider both the temporal and spatial dimensions to obtain accurate and stable results.

\begin{table*}
    \centering
    \resizebox{\textwidth}{!}{
    \begin{tabular}{lllllllllll}
      \toprule
        &\multicolumn{10}{l}{The US dataset}\\
        \cmidrule(l){2-11}
        &\multicolumn{5}{l}{$L$=5}&\multicolumn{5}{l}{$L$=10}\\

        \cmidrule(l){2-6}\cmidrule(l){7-11}
         Model & MAE & RMSE &MAPE & PCC & CCC  & MAE & RMSE &MAPE & PCC & CCC  \\
        \midrule 
        BDGSTN w/o Loss & 3319 & 7547 & 6.18\% &99.80\% & 99.80\%& 5780 & 11984 & 9.27\% &99.55\%  & 99.47\% \\
        BDGSTN w/o Trend &3754  & 8179 & 6.56\% &99.78\% & 99.76\%& 6574 &12778  & 10.50\% &99.54\%  & 99.39\% \\
        BDGSTN & \textbf{3196} & \textbf{7471}& \textbf{6.09\%} & \textbf{99.81\%}& \textbf{99.80\%} & \textbf{5619} & \textbf{11751} & \textbf{9.23\%}& \textbf{99.56\%} & \textbf{99.50\%} \\
        \toprule
        &\multicolumn{5}{l}{$L$=15}&\multicolumn{5}{l}{$L$=20}\\

        \cmidrule(l){2-6}\cmidrule(l){7-11}
          Model& MAE & RMSE &MAPE & PCC & CCC  & MAE & RMSE &MAPE & PCC & CCC  \\
        \midrule 
        BDGSTN w/o Loss & \textbf{7983} & \textbf{15804} & \textbf{13.10\%} &\textbf{99.25\%} & \textbf{99.06\%}&11209  & 21502 & 17.95\% &98.76\%  & 98.20\% \\
        BDGSTN w/o Trend &9609  &  18293& 14.71\% &99.15\% & 98.71\%& 12359 & 22965 & 19.18\% &98.75\%  & 97.91\% \\
        BDGSTN & 8253 & 16193& 13.57\% & 99.22\%& 99.01\% & \textbf{11139} & \textbf{21304} & \textbf{18.06\%}& \textbf{98.78\%} & \textbf{98.23\%} \\
          \bottomrule
    \end{tabular}
    }
    \caption{Ablation study on the US dataset.}
    \label{table6}
\end{table*}
\begin{table*}
    \centering
    \resizebox{\textwidth}{!}{
    \begin{tabular}{lllllllllll}
      \toprule
        &\multicolumn{10}{l}{The Japan Dataset}\\
        \cmidrule(l){2-11}
        &\multicolumn{5}{l}{$L$=5}&\multicolumn{5}{l}{$L$=10}\\

        \cmidrule(l){2-6}\cmidrule(l){7-11}
         Model & MAE & RMSE &MAPE & PCC & CCC  & MAE & RMSE &MAPE & PCC & CCC  \\
        \midrule 
        BDGSTN w/o Loss &996  & 2192 & 18.40\% &98.10\% & 96.18\%& 1492 &3361  & 27.19\% &\textbf{96.38\%}  & 91.90\% \\
        BDGSTN w/o Trend &1042  & 2329 & 18.71\% &97.98\% & 95.73\%& \textbf{1105} & \textbf{2506} & \textbf{21.33\%} &\textbf{96.54\%}  & \textbf{94.80\%} \\
        BDGSTN & \textbf{922} & \textbf{1922}& \textbf{18.07\%} & \textbf{98.34\%}& \textbf{96.93\%} &1407  & 3238 & 25.74\%& 96.27\% & 92.32\% \\
        \toprule
        &\multicolumn{5}{l}{$L$=15}&\multicolumn{5}{l}{$L$=20}\\

        \cmidrule(l){2-6}\cmidrule(l){7-11}
          Model& MAE & RMSE &MAPE & PCC & CCC  & MAE & RMSE &MAPE & PCC & CCC  \\
        \midrule 
        BDGSTN w/o Loss & 1552 & 3739 & 32.24\% &93.63\% & 89.70\%& 1420 & 3428 & 31.98\% &92.18\%  & 90.47\% \\
        BDGSTN w/o Trend & 1313 & 2936 & \textbf{26.95\%} &93.63\% & 92.49\%& 1499 & 3215 & 29.67\% &89.99\%  & 89.46\% \\
        BDGSTN & \textbf{1272} &\textbf{2833} & 26.97\% & \textbf{95.35\%}& \textbf{93.40\%} &  \textbf{1313}& \textbf{2897} & \textbf{28.24\%}& \textbf{92.48\%} & \textbf{92.28\%} \\
          \bottomrule
    \end{tabular}
    }
    \caption{Ablation study on the Japan dataset.}
    \label{table7}
\end{table*}

\subsection{Ablation Study}
\par We further explore the effectiveness of the residual components in the BDGSTN by ablation experiments and propose two variant models: BDGSTN w/o Loss and BDGSTN w/o Trend.
\begin{enumerate}[(1)]
  \item \textbf{BDGSTN w/o Loss}: This model removes the epidemiological module and solely utilizes the spatio-temporal module for epidemic forecasting.
  \item \textbf{BDGSTN w/o Trend}: This model removes the trend and residual components based on time series decomposition in the DLinear model, and directly applies linear transformations to the input features.
\end{enumerate}

\par The results of the ablation experiments on the US and Japan datasets are presented in \reftable{table6} and \reftable{table7}, respectively. By comparing BDGSTN with BDGSTN w/o Loss, BDGSTN achieves better forecasting accuracy in tasks $L$=5, 10, and 20 of the US dataset, as shown in Table 6. Additionally, at $L$=15, BDGSTN also achieves competitive results. On the Japan dataset presented in Table 7, BDGSTN outperforms BDGSTN w/o Loss in all tasks. These results demonstrate the effectiveness of incorporating domain knowledge into the model, facilitating a more precise capture of the fundamental dynamics of epidemics. Compared to BDGSTN w/o Trend, BDGSTN consistently maintains the better performance on the US dataset. On the Japanese dataset, BDGSTN achieves better performance in tasks with $L$=5, 15, and 20. This also reflects the effectiveness of decomposing time series into trend and residual components in DLinear, which can further assist the model in forecasting. Overall, each component in BDGSTN effectively enhances the performance of the model.

\subsection{Comparison of Graph Construction}\label{sub:4.7}
\begin{table*}
    \centering
    \resizebox{\textwidth}{!}{
    \begin{tabular}{lllllllllll}
      \toprule
        &\multicolumn{10}{l}{The US dataset}\\
        \cmidrule(l){2-11}
        &\multicolumn{5}{l}{$L$=5}&\multicolumn{5}{l}{$L$=10}\\

        \cmidrule(l){2-6}\cmidrule(l){7-11}
         Model & MAE & RMSE &MAPE & PCC & CCC  & MAE & RMSE &MAPE & PCC & CCC  \\
        \midrule 
        Geography-based graph  & 7594 &13464  & 10.83\% & 99.38\% & 99.37\% & 8621 & 15316 & 13.11\% & 99.18\% & 99.16\%   \\
        Gravity-based graph  & 5795 & 11644 & 9.36\% & 99.69\% & 99.55\% & 6727 & 13153 & 11.89\% & 99.47\% &  99.41\%  \\
        DTW-based graph  & 13392 & 22725 & 16.09\% & 99.45\% & 97.96\% & 15173 & 25866 & 17.82\% & 99.23\% &  97.29\%  \\
        PCC-based graph  & 5051 & 10142 & 10.92\% & 99.68\% & 99.63\% &  6591 & 13151 & 13.56\% & 99.46\% &  99.37\%  \\
        Backbone graph  & 3443 & 7830 & \textbf{6.00\%} & 99.80\% & 99.78\% & 6120 & 12388 & 9.56\% & 99.56\% & 99.43\%   \\
        Temporal graph  & 4692 & 9124 & 8.02\% & 99.71\% & 99.70\% & 5769 & \textbf{11401} & 10.06\% & 99.56\% & \textbf{99.53\%}   \\
        Backbone-based graph  & \textbf{3196} & \textbf{7471} & 6.10\% & \textbf{99.81\%} & \textbf{99.80\%} & \textbf{5619} & 11751 &  \textbf{9.24\%}& \textbf{99.56\%} &  99.50\%  \\
        \toprule
        &\multicolumn{5}{l}{$L$=15}&\multicolumn{5}{l}{$L$=20}\\

        \cmidrule(l){2-6}\cmidrule(l){7-11}
          Model& MAE & RMSE &MAPE & PCC & CCC  & MAE & RMSE &MAPE & PCC & CCC  \\
        \midrule 
        Geography-based graph  & 10313 & 18219 & 16.41\% & 98.86\% & 98.78\% & 12851 & 22779 & 20.71\% & 98.36\% & 98.02\%   \\
        Gravity-based graph  & 8334 & \textbf{15740} & 15.36\% & 99.14\% & \textbf{99.13\%} & \textbf{10944} &\textbf{19559}  & 19.76\% & 98.65\% & \textbf{98.60\%}   \\
        DTW-based graph  & 18510 & 32059 & 21.33\% & 98.67\% & 95.67\% & 22862 & 39321 & 25.92\% & 98.11\% &  93.23\%  \\
        PCC-based graph  & 8934 & 17353 & 17.32\% & 99.10\% & 98.87\% & 12260 & 23151 & 21.96\% & 98.54\% &  97.91\%  \\
        Backbone graph  & 9213 & 17955 & 13.91\% & 99.19\% & 98.76\% & 12236 & 23621 & 18.70\% &98.67\%  &  97.78\%  \\
        Temporal graph  & 8791 & 16538 & 14.55\% & 99.03\% & 98.97\% & 12487 & 22574 & 19.76\% & 98.34\% & 98.05\%   \\
        Backbone-based graph  & \textbf{8253} & 16193 & \textbf{13.57\%} & \textbf{99.22\%} & 99.01\% &11139  &  21304&\textbf{18.06\%}  & \textbf{98.78\%} & 98.23\%   \\
          \bottomrule
    \end{tabular}
    }
    \caption{Comparison of graph construction on the US dataset.}
    \label{table8}
\end{table*}
\begin{table*}
    \centering
    \resizebox{\textwidth}{!}{
    \begin{tabular}{lllllllllll}
      \toprule
        &\multicolumn{10}{l}{The Japan Dataset}\\
        \cmidrule(l){2-11}
        &\multicolumn{5}{l}{$L$=5}&\multicolumn{5}{l}{$L$=10}\\
        \cmidrule(l){2-6}\cmidrule(l){7-11}
         Model & MAE & RMSE &MAPE & PCC & CCC  & MAE & RMSE &MAPE & PCC & CCC  \\
        \midrule 
        Geography-based graph  & 1023 & 2187 & 19.73\% & 97.95\% & 96.17\% & \textbf{1400} & \textbf{3098} & 27.06\% & 96.37\% & \textbf{92.81\%}   \\
        Gravity-based graph  & 944 & 2119 & \textbf{14.32\%} & \textbf{98.48\%} & 96.48\% & \textbf{1253} & \textbf{2888} & \textbf{20.73\%} & 96.83\% &  \textbf{93.67\%}  \\
        DTW-based graph  & 1170 & 2576 & 19.82\% & 97.15\% & 94.85\% & 1448 & \textbf{3173} & \textbf{25.25\%} & 96.33\% & \textbf{92.53\%}   \\
        PCC-based graph  & 1089 & 2288 & 18.12\% & 97.53\% & 95.84\% &  1442& \textbf{3186} & \textbf{25.29\%} & 95.80\% & \textbf{92.44\% }  \\
        Backbone graph  & 952 & 2093 & \textbf{17.16\%} & 98.34\% & 96.51\% &\textbf{1317}  & \textbf{3059} &\textbf{23.96\%}  & 96.55\% &  \textbf{93.02\%}  \\
        Temporal graph  & 1044 & 2290 & 20.28\% & 97.90\% & 95.82\% & 1423 & \textbf{3056} & 27.55\% & 96.45\% & \textbf{92.86\%}   \\
        Backbone-based graph  & \textbf{922} & \textbf{1922} & 18.07\% & 98.34\% & \textbf{96.93\%} & 1407 & 3238 & 25.74\% & 96.27\% & 92.32\%   \\
        \toprule
        &\multicolumn{5}{l}{$L$=15}&\multicolumn{5}{l}{$L$=20}\\

        \cmidrule(l){2-6}\cmidrule(l){7-11}
          Model& MAE & RMSE &MAPE & PCC & CCC  & MAE & RMSE &MAPE & PCC & CCC  \\
        \midrule 
        Geography-based graph  & 1466 &3316  & 31.85\% & 94.32\% & 91.45\% & 1394 & 3221 & 31.94\% & 91.89\% & 91.05\%   \\
        Gravity-based graph  & 1399 & 3291 & \textbf{26.56\%} & 93.80\% & 91.45\% & 1474 & 3279 & 29.25\% & 89.90\% & 89.77\%   \\
        DTW-based graph  & 1550 & 3409 & 29.96\% & 94.57\% & 91.16\% & 1449 & 3115 & 29.44\% & 92.31\% & 91.54\%   \\
        PCC-based graph  & 1596 & 3656 & 31.55\% & 92.58\% & 89.76\% & 1717 & 4021 & 36.08\% & 87.16\% & 86.31\%   \\
        Backbone graph  & 1405 & 3352 & 29.26\% & 94.39\% & 91.38\% & 1352 & 3176 & 30.17\% & 91.86\% & 91.22\%   \\
        Temporal graph  & 1582 & 3564 & 33.37\% & 92.26\% & 89.77\% & 1681 & 4283 & 36.65\% & 89.32\% & 86.19\%   \\
        Backbone-based graph  & \textbf{1272} &\textbf{2833} & 26.97\% & \textbf{95.35\%}& \textbf{93.40\%} &  \textbf{1313}& \textbf{2897} & \textbf{28.24\%}& \textbf{92.48\%} & \textbf{92.28\%} \\
          \bottomrule
    \end{tabular}
    }
    \caption{Comparison of graph construction on the Japan dataset.}
    \label{table9}
\end{table*}

\begin{figure*}[t]
	\centering  
	\subfigure[Graph structure at T=1]{
		\includegraphics[width=0.325\linewidth]{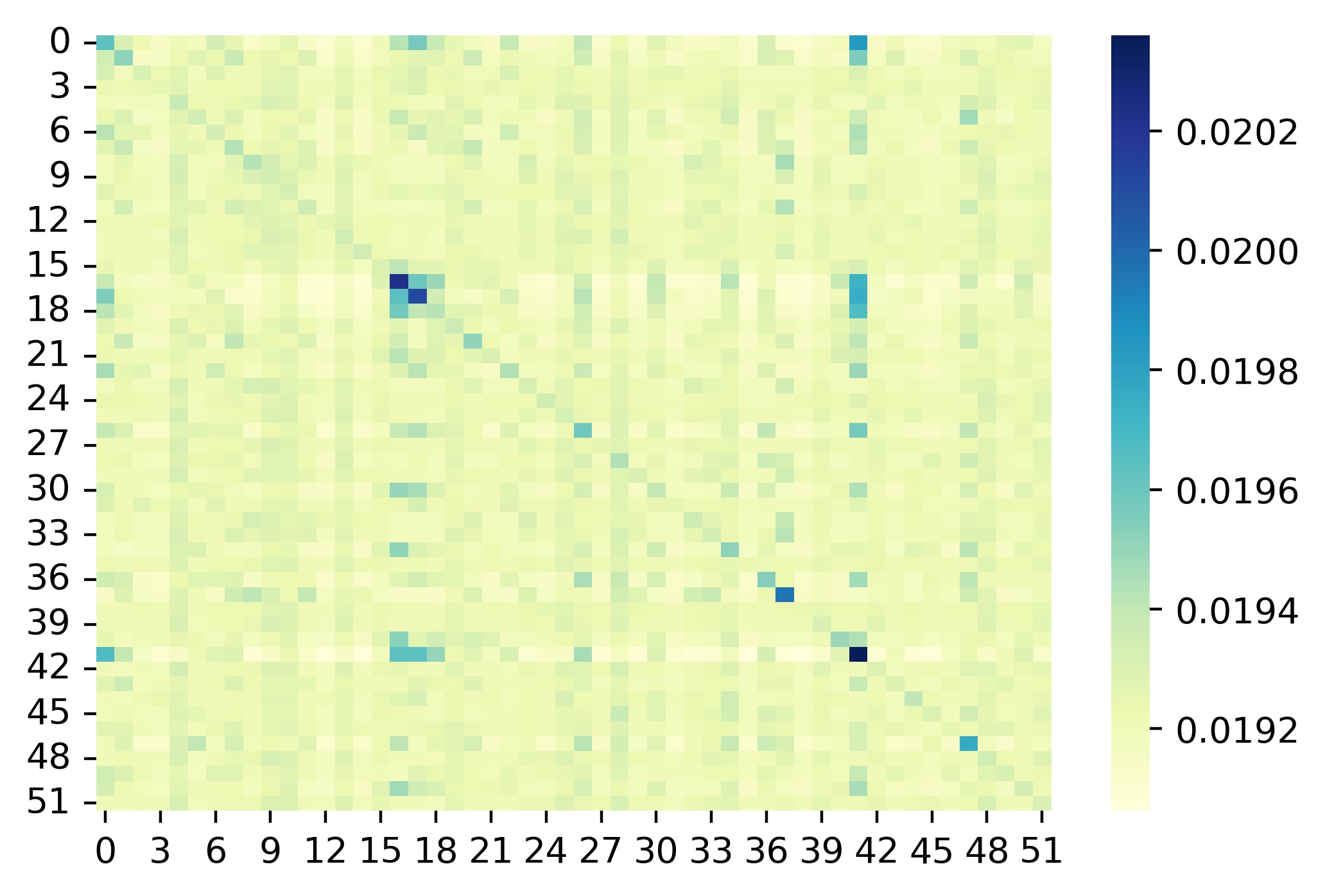}}
        \subfigure[Graph structure at T=5]{
		\includegraphics[width=0.325\linewidth]{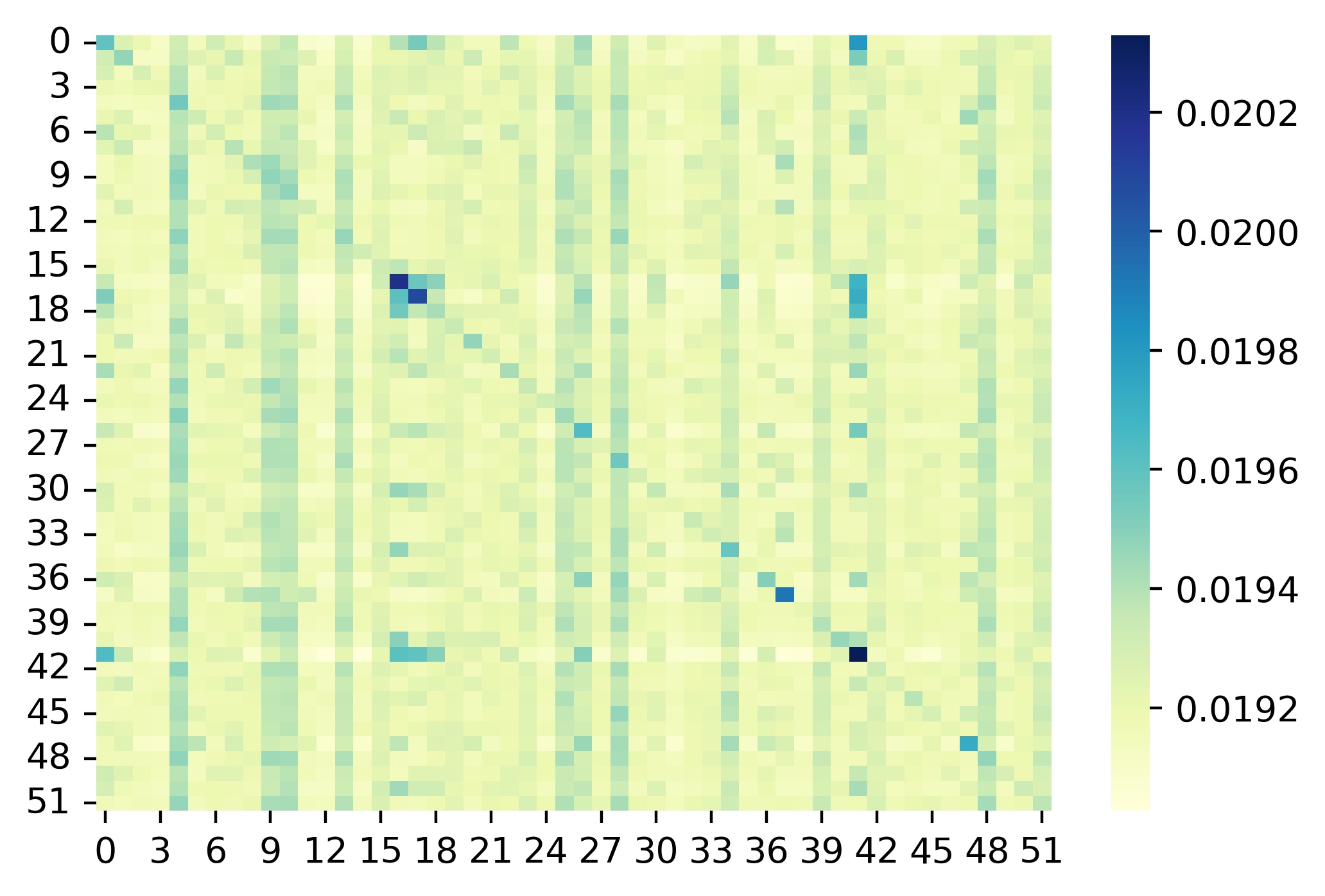}}
        \subfigure[Differences between graph structures at T=1 and T=5]{
		\includegraphics[width=0.325\linewidth]{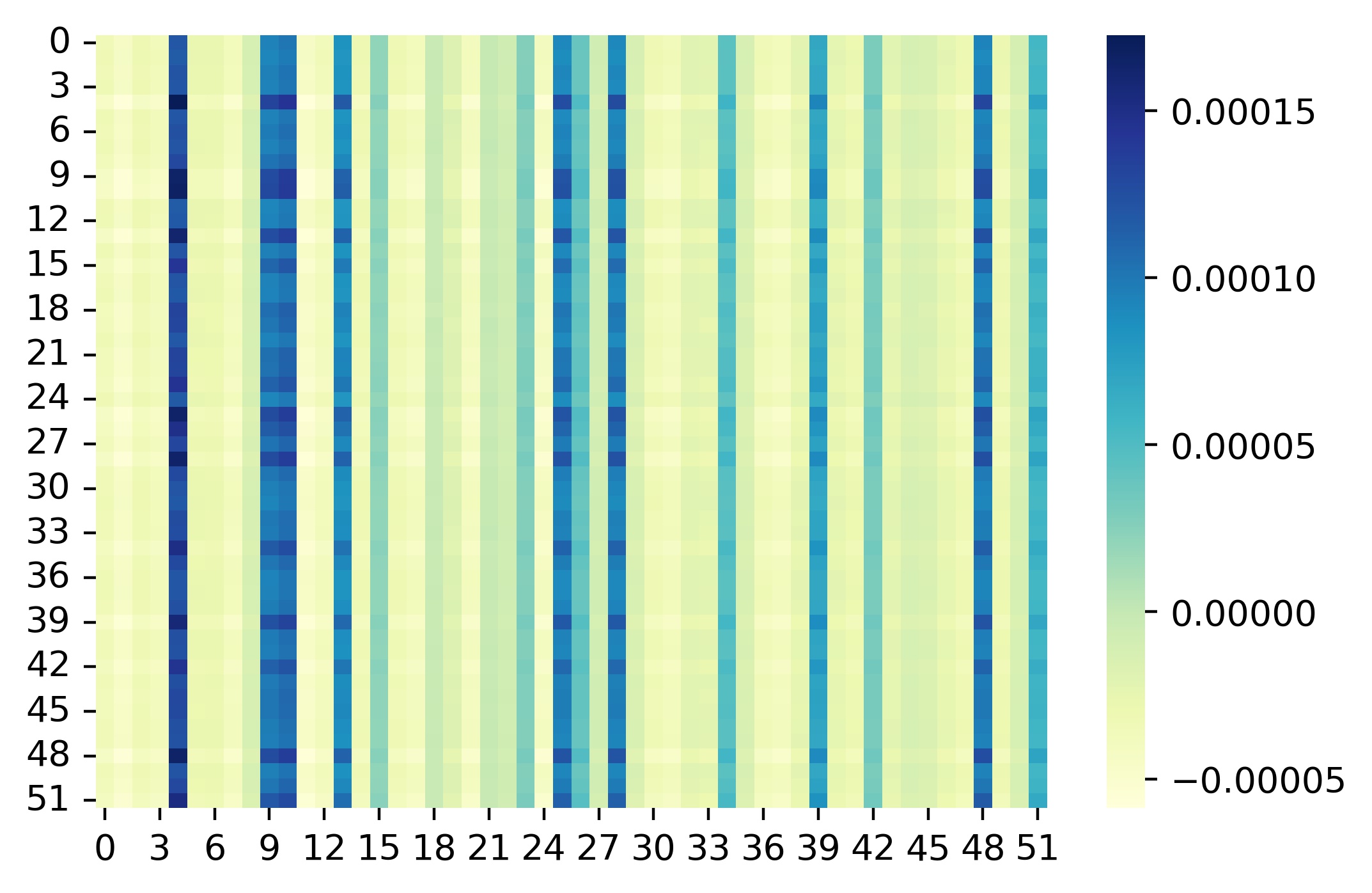}}
	\caption{Visualization of learned graph structures.}
    \label{fig4}
\end{figure*}

\par The rationality and effectiveness of graph construction methods have a significant impact on spatial-level information aggregation. To validate the superiority of our proposed backbone-based dynamic graph generation method, we compare it against six different graph construction methods. These comparative methods can be categorized into three groups: (\romannumeral1) prior knowledge: Geography-based graph and Gravity-based graph, (\romannumeral2) data similarity: DTW-based graph and PCC-based graph, and (\romannumeral3) learning-based generation: Static graph, and Dynamic graph.

\begin{enumerate}[(1)]
  \item \textbf{Geography-based graph}: Constructing a graph structure based on the geographic adjacency between patches in the real world. 
  \item \textbf{Gravity-based graph}: Constructing a graph structure by combining the distance between populations and their population sizes using a gravity model.
  \item \textbf{DTW-based graph}: Constructing a graph structure by applying the Dynamic Time Warping (DTW) algorithm to calculate the similarity between patch time series.
  \item \textbf{PCC-based graph}: Constructing a graph structure by applying the Pearson’s Correlation Coefficient (PCC) algorithm to calculate the similarity between patch time series.
  \item \textbf{Backbone graph}: The temporal graph component is removed from the proposed method, and only learnable embeddings are used to generate a static graph.
  \item \textbf{Temporal graph}: The backbone graph component is removed from the proposed method, and only TCN is used to generate a dynamic graph.
\end{enumerate}

\par The comparison results on the US and Japan datasets are presented in \reftable{table8} and \reftable{table9}. Based on the experimental results, it is evident that our proposed backbone-based dynamic graph structure achieves the best prediction accuracy across multiple tasks and different datasets. When compared to graph structures based on prior knowledge, it is clear that the Geography-based graph yields inferior results. This is because the graph structure constructed using prior knowledge fails to fully capture the information within the graph, including the potential interactions between non-adjacent patches. Moreover, the Gravity-based graph exhibits better forecasting performance than the Geography-based graph and even outperforms our proposed dynamic graph method in certain tasks (e.g., Japan dataset with $L$=10). This phenomenon can be attributed to the fact that the gravity model incorporates additional prior knowledge, such as the distance between patches and their population sizes, thereby facilitating the construction of a more comprehensive graph structure.

\par In comparison to methods that rely on data similarity to generate graph structures, we observe that both DTW-based graph and PCC-based graph fall short in achieving more accurate epidemic forecasting. Specifically, the DTW-based graph produces notably poor forecasting results in the US dataset, as demonstrated in Table 8. Consequently, we believe that relying solely on data similarity for constructing graph structures is not conducive to effective epidemic forecasting.

\par In the context of learning-based generation methods, our focus is primarily on investigating the effectiveness of the backbone-based dynamic graph structure. By comparing it with models that solely rely on backbone graph structures or temporal graph structures, we observe that incorporating the static graph structure as the backbone and integrating dynamic temporal graph information leads to improved forecasting accuracy. This finding suggests that relying solely on either static or dynamic information is inadequate, thereby underscoring the effectiveness of our proposed dynamic graph construction method.

\par To understand the changes in the dynamic graph structures, we present the graph structures learned by BDGSTN at $T$=1 and 5 in the US dataset in \reffig{fig4}(a)(b), and highlight the differences in graph information between $T$=1 and $T$=5 in \reffig{fig4}(c). It can be observed that the information between graph structures at different time steps is similar, while \reffig{fig4}(c) shows variations across different time steps. This is due to the fact that our proposed graph learning method integrates backbone graph information with temporal graph information to construct dynamic graphs. Furthermore, we find that the graph at different time steps exhibits a higher influence along the diagonal, which is reasonable as patches should have a significant impact on themselves. Additionally, the graph reveals that certain patches exert a higher influence on others, potentially representing patches with greater real-world influence.

\begin{figure*}[t]
	\centering  
	\subfigure[US]{
		\includegraphics[width=0.49\linewidth]{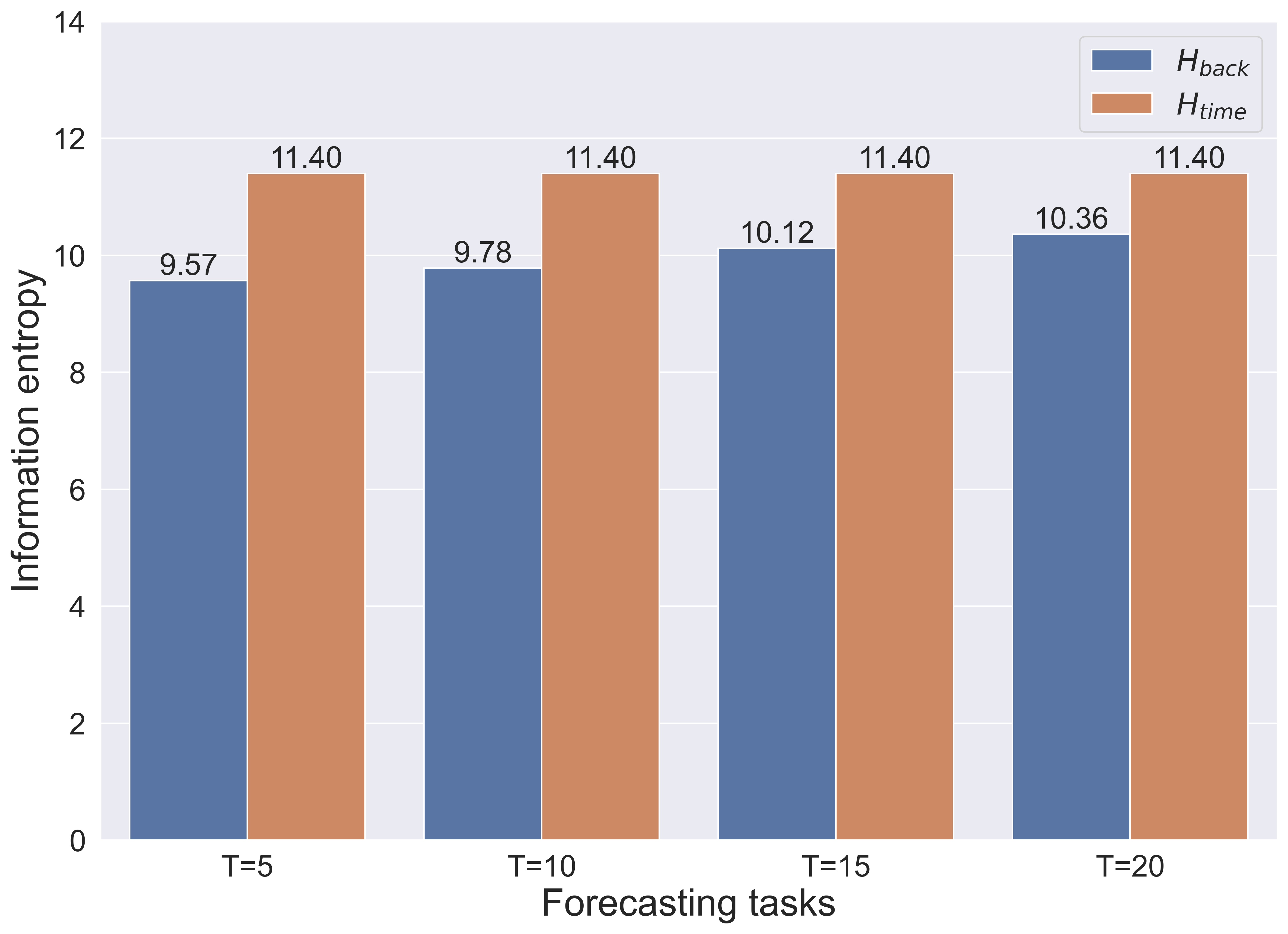}}
	\subfigure[Japan]{
		\includegraphics[width=0.49\linewidth]{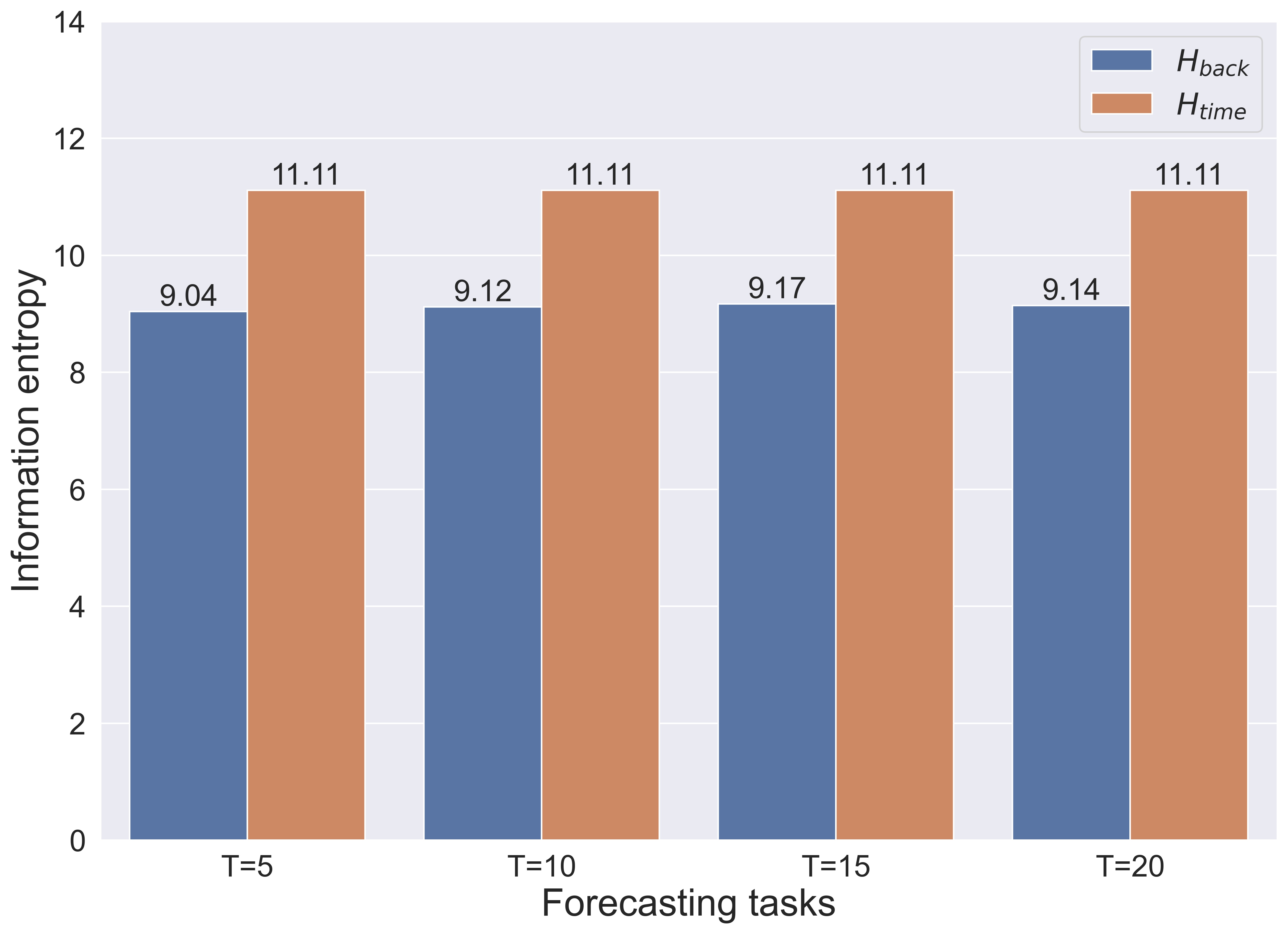}}
        
	\caption{Comparison of information entropy in US(a) and Japan(b).}
    \label{fig5}
\end{figure*}

\begin{figure*}[t]
	\centering  
	\subfigure[US]{
		\includegraphics[width=0.49\linewidth]{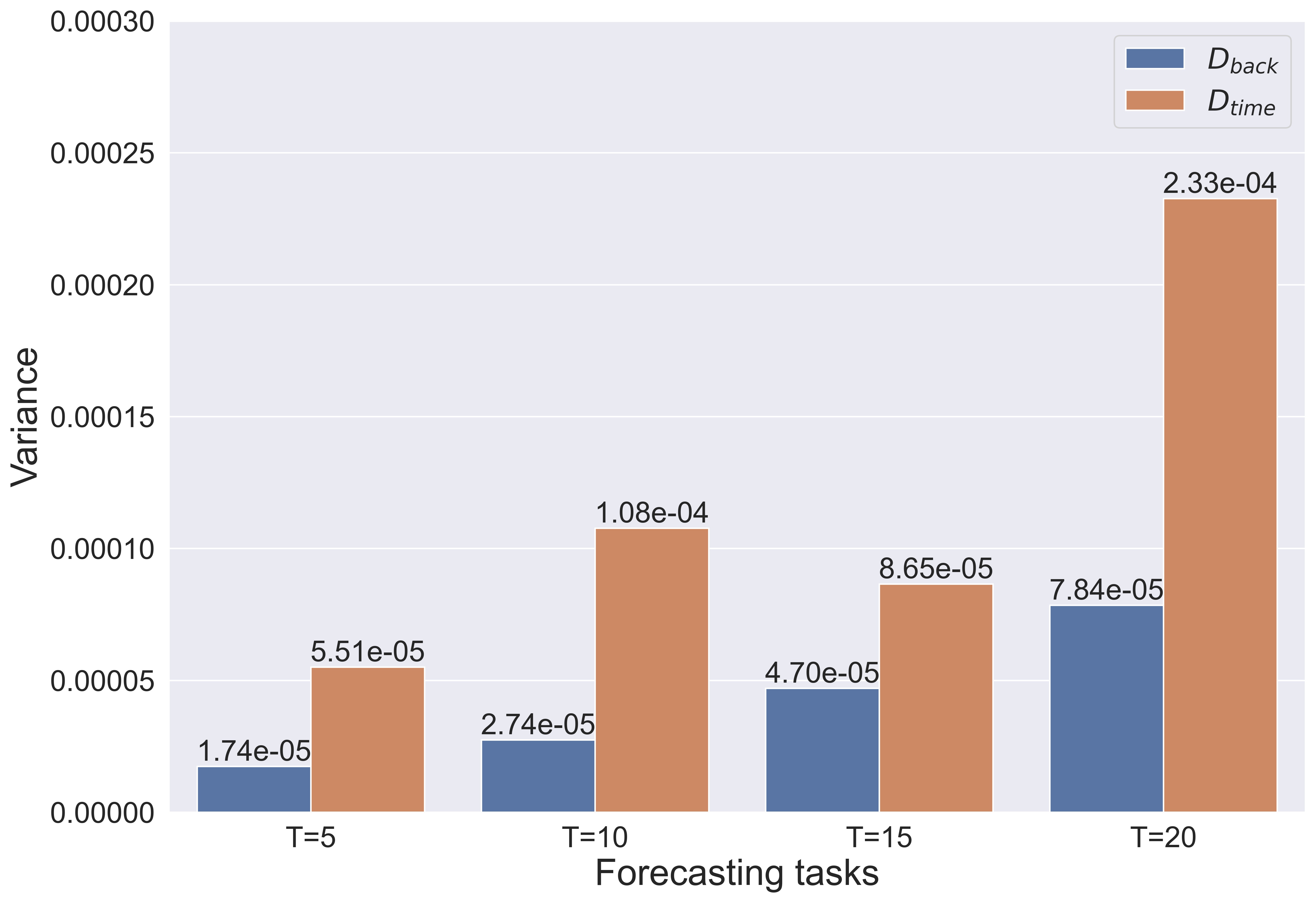}}
	\subfigure[Japan]{
		\includegraphics[width=0.49\linewidth]{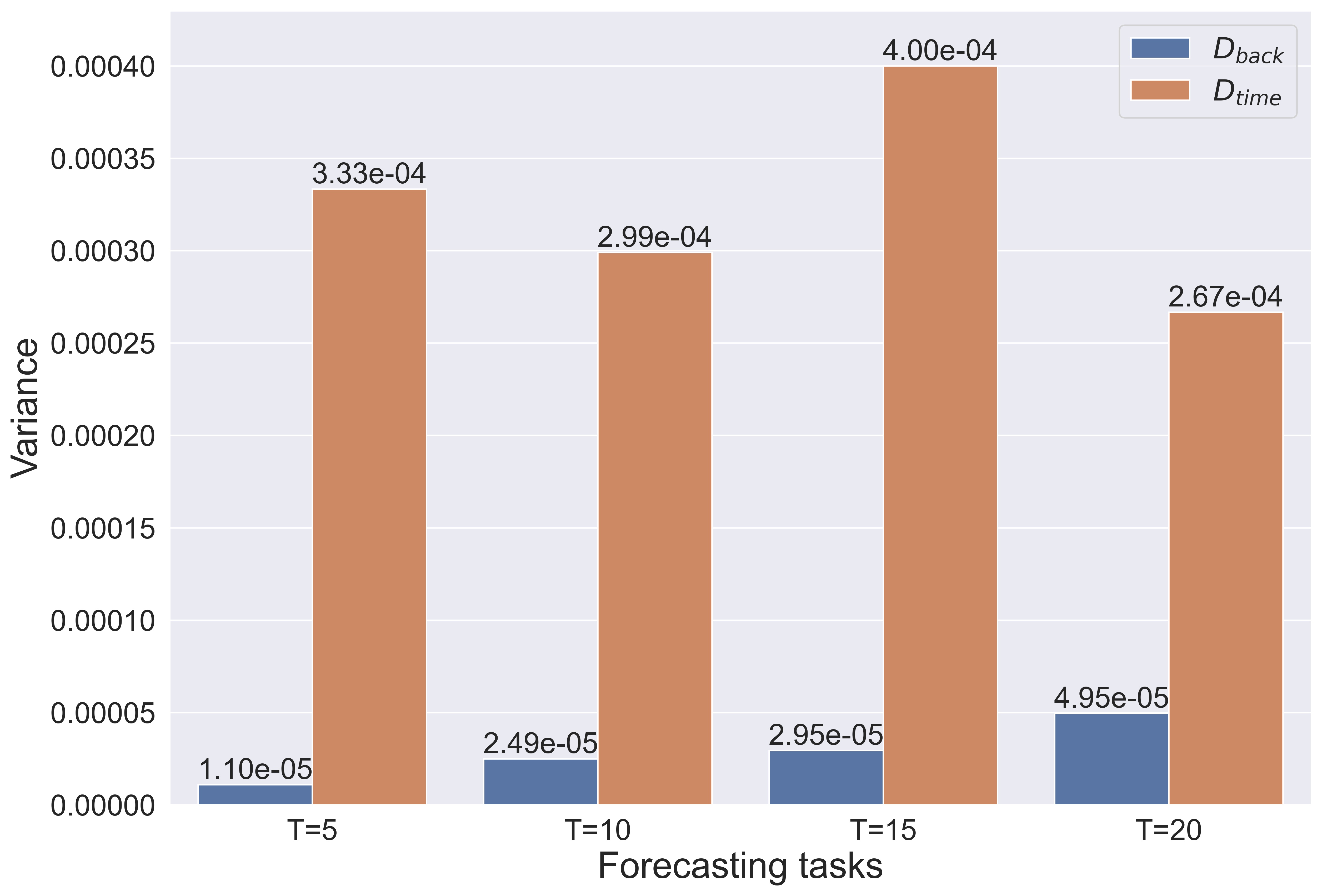}}
        
	\caption{Comparison of variance in US(a) and Japan(b).}
    \label{fig6}
\end{figure*}

\begin{figure*}[t]
	\centering  
	\subfigure[US]{
		\includegraphics[width=0.49\linewidth]{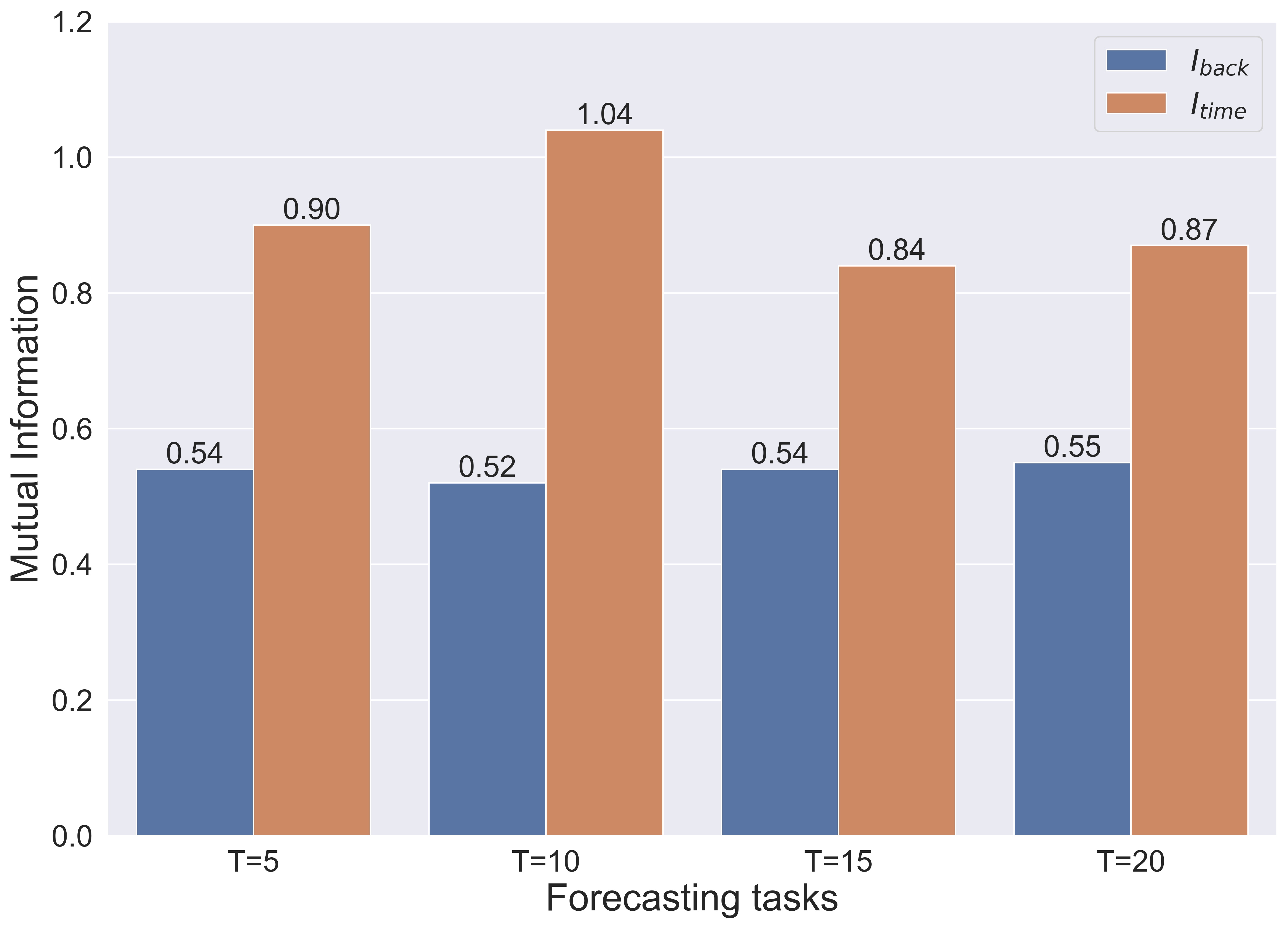}}
	\subfigure[Japan]{
		\includegraphics[width=0.49\linewidth]{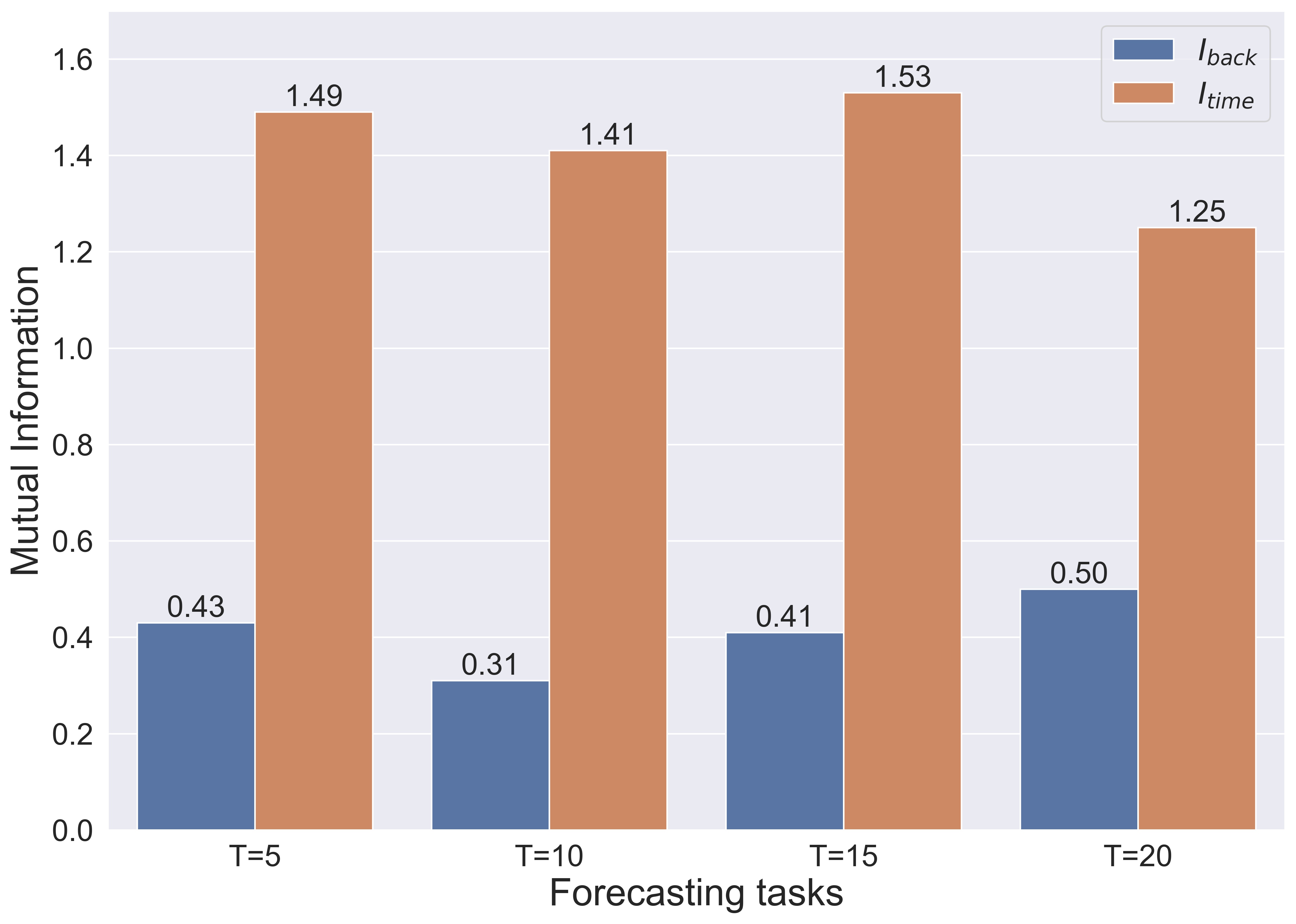}}
        
	\caption{Comparison of mutual information in US(a) and Japan(b).}
    \label{fig7}
\end{figure*}

\subsection{Dynamic Graph Discussion}\label{sub:4.8}
\par To further analyze and discuss the respective roles of the backbone graph and the temporal graph in dynamic graphs, we conduct an analysis with information metrics on them. We select information entropy and variance as metrics to assess the stability and flexibility of backbone and temporal graphs, and employ mutual information to quantify their correlation with the generated dynamic graph.
\par Information entropy is a metric used to quantify the uncertainty of a random variable~\cite{ma2018shannon}. It can also be applied to measure the stability and diversity of the edge weight distribution in backbone and temporal graphs. A higher information entropy indicates a wider distribution, reflecting diversity in edge weights, while a lower information entropy suggests a more concentrated distribution, reflecting stability in edge weights. We firstly discretize the weight matrices of the backbone and temporal graphs, and then calculate their information entropy separately as follows:
\begin{align}
    &H_{back}=-\sum_{i=1}^{N_{back}}p_{i}\log_{2}{p_{i}},\\
    &H_{time}=-\sum_{j=1}^{N_{time}}p_{j}\log_{2}{p_{j}},
\end{align}%
where $H_{back}$ and $H_{time}$ represent the information entropy of the backbone graph and temporal graph, respectively, $N_{back}$ and $N_{time}$ represent the number of categories for the edge weights in the backbone graph and temporal graph, respectively, and $p_{i}$ represents the probability of an edge weight being in the $i$-th category.

\par The experimental results on the information entropy from the US and Japan datasets are shown in \reffig{fig5}. It is evident that the entropy of the backbone graph $H_{back}$ is consistently lower than that of the temporal graph $H_{time}$ across all tasks. This observation implies that the generated backbone graph is relatively more stable and deterministic, effectively capturing the primary patterns or structures in the data. Consequently, the backbone graph yields more stable outputs, reflecting fixed features or patterns within the dataset. Conversely, the temporal graph displays a higher entropy, indicating its greater flexibility and diversity. The temporal graph captures dynamic information about the temporal sequence, introducing more uncertainty and diversity. This flexibility allows the temporal graph to adapt to various changes, providing more flexible and diverse outputs.

\par Using discrete information entropy as an approximation method to measure uncertainty is not capable of accurately capturing the level of uncertainty in continuous values. Therefore, we also employ variance as a metric to assess the level of uncertainty in the generated graph, and the results are shown in \reffig{fig6}. A higher variance signifies larger differences between the edge weights, indicating more dispersion or uncertainty. Conversely, a lower variance suggests smaller differences between the edge weights, indicating more concentration or certainty. The calculation is as follows:
\begin{align}
    &D(X)=E\left [ (X-E(X))^{2} \right ],
\end{align}%
where $X$ represents edge weight matrix, $E(X)$ represents the mean value of $X$, and $D(X)$ represents the variance.

\par By observing \reffig{fig6}, it is apparent that the variance of the backbone graph $D_{back}$ consistently remains lower than that of the temporal graph $D_{time}$ across all tasks. This experimental result aligns with the findings from the information entropy experiments, indicating that the backbone graph is more stable and deterministic, while the temporal graph is more uncertain and diverse. Based on this differentiated learning approach, we separately learn the backbone graph and the temporal graph. By merging these two generated graphs, we create a dynamic graph that combines the stability and determinism of the backbone graph with the flexibility and diversity of the temporal graph. This fusion approach enhances the expressiveness and adaptability of the dynamic graph.

\par After discussing the backbone and temporal graphs separately, we further investigate their correlation with the generated dynamic graph. Mutual information is a measure of correlation between two random variables~\cite{batina2011mutual}. A higher mutual information value indicates a stronger correlation between the two random variables, while a lower mutual information value indicates a weaker correlation. To understand the correlation between them, we use a KNN-based entropy estimation method~\cite{kraskov2004estimating} to calculate the following mutual information:
\begin{align}
    &I(X,Y)=\Psi (k)-\left \langle \Psi (n_x+1)+\Psi (n_y+1) \right \rangle+\Psi(N),
\end{align}%
where $I(X,Y)$ represents the mutual information between random variables $X$ (backbone graph or temporal graph) and $Y$ (dynamic graph), $k$ represents the number of neighbors, $\Psi$ represents the double gamma function, $n_x$ and $n_y$ represent the number of data pairs within the radius determined by the KNN algorithm, $N$ represents the total number of samples, and $<>$ denotes the averaging operation.

\par The experimental results, as shown in \reffig{fig7}, reveal that the mutual information between the backbone graph and the dynamic graph $I_{back}$ is consistently lower than that between the temporal graph and the dynamic graph $I_{time}$ across all tasks. This suggests that $I_{time}$ exhibits a stronger correlation than $I_{back}$, implying that the temporal graph plays a more prominent role in generating the fused dynamic graph, and its features are better reflected in the variations of the dynamic graph. This is reasonable as the fused dynamic graph displays more flexibility and diversity of the temporal graph, while the backbone graph mainly learns the main structure of the dynamic graph. Therefore, the stability and determinism of the backbone graph are less evident in the generated dynamic graph. This also indicates that dynamic information is crucial in generating graph structures.

\subsection{Model Complexity and Efficiency}
\par
\begin{figure*}[t]
\centering
\includegraphics[width=0.8\linewidth]{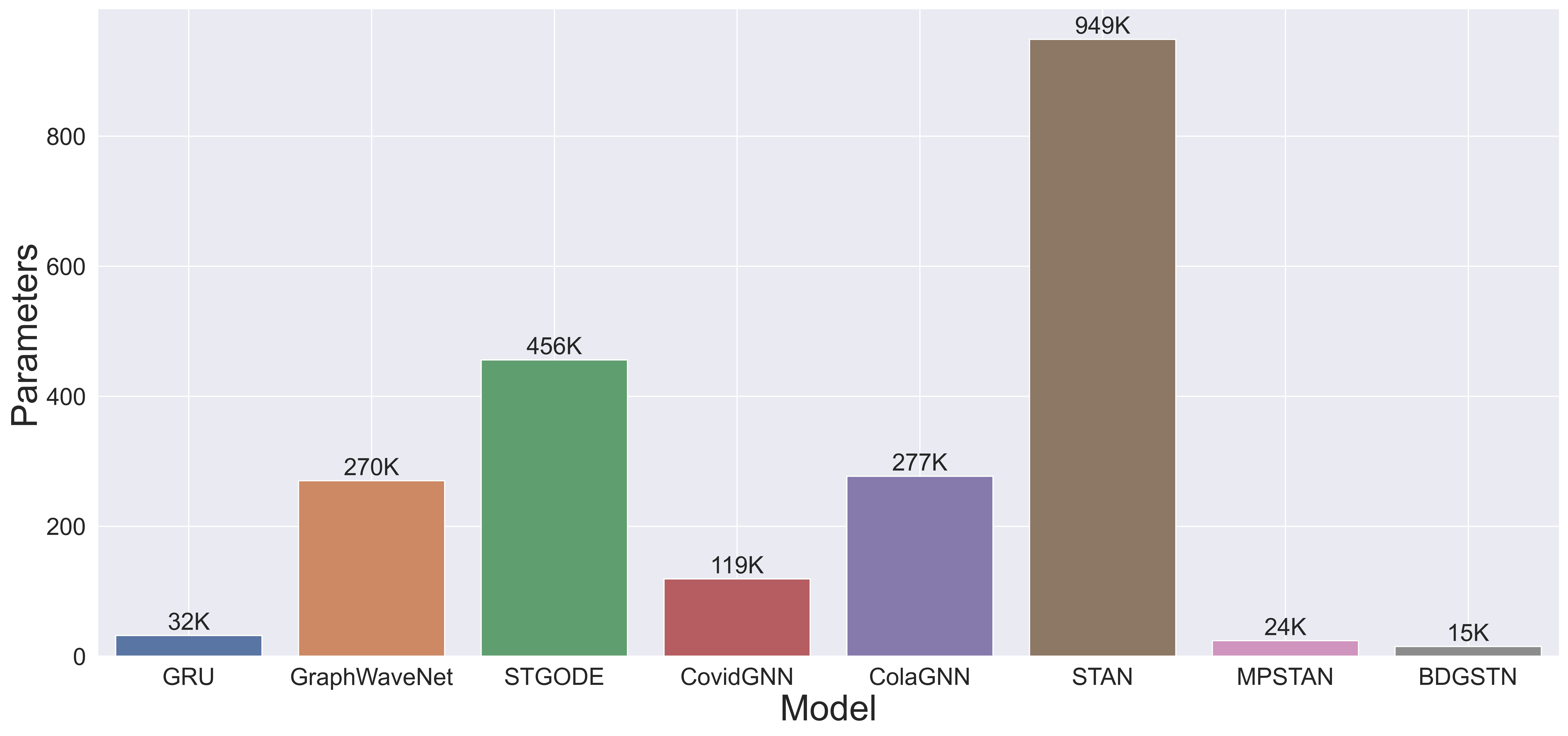}
\caption{Comparison of model complexity.}
\label{fig8}
\end{figure*}

 We analyze model complexity by comparing the neural network parameter volume of BDGSTN with that of other deep learning models. As shown in \reffig{fig8}, it is evident that BDGSTN has the lowest parameter volume. This is attributed to our simple spatio-temporal epidemic forecasting framework, which primarily employs a linear transformation model to capture temporal dependencies. Furthermore, we compare BDGSTN's training time with the advanced spatio-temporal epidemic model MPSTAN's one, as shown in \reftable{table10} for the US dataset. We observe that BDGSTN outperforms MPSTAN in terms of time consumption. The efficiency disadvantage of MPSTAN stems from its recurrent structure, which involves extensive inter-patch calculations, resulting in lower efficiency. In addition, as the forecasting window increases, the training time also experiences a significant rise. However, BDGSTN utilizes a simpler structure that achieves improved forecasting accuracy while reducing computational requirements, making the corresponding increase in training time acceptable.

\section{Conclusion} \label{sec:Conclusion}
\par In this paper, we propose a backbone-based dynamic graph spatio-temporal network (BDGSTN) for epidemic forecasting. Our model introduces a novel approach to generate the dynamic graph by combining backbone graph learning and temporal graph learning. By integrating the dynamic graph, DLinear, GCN, and SIR models into a simple spatio-temporal epidemic framework, we aim to reduce computational consumption and improve efficiency. Experimental results demonstrate that BDGSTN outperforms other baseline models on two different datasets, and ablation comparison verifies the effectiveness of each component of the model. Additionally, we compare different graph generation methods and validate that the backbone-based dynamic graph yields more stable and accurate forecasting. Furthermore, we employ information metrics to discuss the roles of the backbone graph and the temporal graph in dynamic graph generation.  Finally, the comparison of model parameter volume and training time verifies the superiority of BDGSTN in terms of model complexity and efficiency.
\par In the future, we will explore more optimal fusion methods for the backbone graph and the temporal graph, such as gate mechanisms or attention weights, to generate dynamic graphs with enhanced expressive power. Moreover, we will conduct research on methods to better integrate epidemiological knowledge into neural networks, aiming to assist the model in capturing underlying epidemic dynamics more effectively without significantly increasing training time.

\begin{table*}
    \centering
    \resizebox{0.7\textwidth}{!}{
    \begin{tabular}{lllllllllll}
      
        \toprule
        &\multicolumn{4}{l}{The US Dataset}\\
        \cmidrule(l){2-5}
        &\multicolumn{2}{l}{$L$=5}&\multicolumn{2}{l}{$L$=10}\\

        \cmidrule(l){2-3}\cmidrule(l){4-5}
          Model& Per epoch time (s) & Training time (s)& Per epoch time (s)& Training time (s)\\
        \midrule 
        MPSTAN  & 14.7 & 735 & 21.34 & 1067  \\
        BDGSTN  & 0.285 & 57 & 0.375 & 75  \\
          \toprule
        &\multicolumn{2}{l}{$L$=15}&\multicolumn{2}{l}{$L$=20}\\
        \cmidrule(l){2-3}\cmidrule(l){4-5}
          Model& Per epoch time (s)& Training time (s)& Per epoch time (s)& Training time (s)\\
        \midrule 
        MPSTAN  & 27.26 & 1363 & 33.6 & 1680  \\
        BDGSTN  & 0.435 & 87 & 0.495 & 99  \\
          \bottomrule
    \end{tabular}
    }
    \caption{Comparison of model efficiency.}
    \label{table10}
\end{table*}





\printcredits
\vskip 0.7 cm\noindent\textbf{Declaration of competing interest}

The authors declare that they have no known competing financial interests or personal relationships that could have appeared to influence the work reported in this paper.

\vskip 0.7 cm\noindent\textbf{Acknowledgments}

This work was supported by the National Natural Science Foundation of China (Grant No. 52273228), Natural Science Foundation
of Shanghai, China(Grant No. 20ZR1419000), Key Research Project of
Zhejiang Laboratory (Grant No. 2021PE0AC02).
\bibliographystyle{model1-num-names}
\bibliography{references}
\end{document}